\RequirePackage{fix-cm}
\documentclass{svjour3} 
\usepackage[usestackEOL]{stackengine}
\usepackage{adjustbox}
\usepackage{floatflt}
\usepackage{caption}
\usepackage{subcaption}
\usepackage{multirow}
\usepackage{cite}
\usepackage{amsmath,amssymb,amsfonts}
\usepackage{algorithmic}
\usepackage{graphicx,url}
\usepackage{textcomp}
\usepackage{xcolor}

\usepackage[utf8x]{inputenc}
\graphicspath{{figs/}}
\DeclareMathOperator*{\argmax}{arg\,max}
\DeclareMathOperator*{\argmin}{arg\,min}

\begin{document}

\title{Towards Trustworthy AutoGrading of Short, Multi-lingual, Multi-type Answers}  
\subtitle{}
\author{Johannes Schneider \and Robin Richner \and Micha Riser }
\institute{J. Schneider \at Institute of Information Systems\\ University of Liechtenstein, Liechtenstein \\
\email{johannes.schneider@uni.li}
\and R. Richner \at Classtime / Institute of Information Systems \\ University of Liechtenstein, Liechtenstein   \\
\email{robin.richner@uni.li}
\and M. Riser \at Classtime  \\ Zurich, Switzerland  \\
\email{micha@classtime.com}
}

%{ \large Subtitle as needed \textbf{\textit{(paper subtitle)}} \vspace{2em}}
%\thanks{\hrule height 0.5pt \vspace{4pt} Identify applicable sponsor/s here. If no sponsors, delete this text box (sponsors).}
% \author{\IEEEauthorblockN{Johannes Schneider}
% \IEEEauthorblockA{University of Liechtenstein \\
% johannes.schneider@uni.li}
% \and
% \IEEEauthorblockN{Robin Richner}
% \IEEEauthorblockA{Classtime / University of Liechtenstein \\
% robin.richner@protonmail.ch} \\
% \and
% \IEEEauthorblockN{Micha Riser}
% \IEEEauthorblockA{Classtime \\
% micha@classtime.com}
% }
\date{Received: date / Accepted: date}
\maketitle

%inlcuding state-of-the-art models such as BERT have been used,
%while for such cases impressive results have been shown,
\begin{abstract}
Autograding short textual answers has become much more feasible due to the rise of NLP and the increased availability of question-answer pairs brought about by a shift to online education. Autograding performance is still inferior to human grading. The statistical and black-box nature of state-of-the-art machine learning models makes them untrustworthy, raising ethical concerns and limiting their practical utility. Furthermore, the evaluation of autograding is typically confined to small, monolingual datasets for a specific question type. 
This study uses a large dataset consisting of about 10 million question-answer pairs from multiple languages covering diverse fields such as math and language, and strong variation in question and answer syntax. We demonstrate the effectiveness of fine-tuning transformer models for autograding for such complex datasets. Our best hyperparameter-tuned model yields an accuracy of about 86.5\%, comparable to the state-of-the-art models that are less general and more tuned to a specific type of question, subject, and language.
More importantly, we address trust and ethical concerns. By involving humans in the autograding process, we show how to improve the accuracy of automatically graded answers, achieving accuracy equivalent to that of teaching assistants. We also show how teachers can effectively control the type of errors made by the system and how they can validate efficiently that the autograder's performance on individual exams is close to the expected performance. %Our work is among the first that (i) effectively discusses trust issues and possible solutions arising from imperfect decisions of autograders, (ii) allows to take into account a teacher’s preferences for errors.%, and (iii) discusses and allows to address ethical considerations with respect to errors in answers. 

%We discuss a simple mechanism so that the ``autograder'' only makes a decision if it is very certain that its decision is indeed correct (otherwise it is deferred to a human assessor). The resulting accuracy for classified samples is roughly comparable to teaching assistants, while still yielding a significant overall workload reduction. The data labeled by the teacher is also used to verify the autograder's performance for a particular with respect to deviation of performance as obtained by a large test set. This mechanism is among the first that (i) effectively discusses trust issues and possible solutions arising from imperfect decisions of autograders, (ii) allows to take into account a teacher’s preferences, and (iii) discusses and allows to address ethical considerations with respect to errors in answers. 

\keywords{Autograding \and  NLP \and multi-lingual \and BERT \and question-answering \and validation}
\end{abstract}

%\begin{IEEEkeywords}
%
%\end{IEEEkeywords}

\section{Introduction} %While it seems that fully automated marking of arbitrary textual answers for complex questions is unrealistic any time soon, a There are millions of teachers worldwide, even in a small country such as Switzerland there are more than 120000 teachers\footnote{\url{https://en.wikipedia.org/wiki/Education_in_Switzerland}} e.g.
% \cite{Martin2014}
Teachers and professors spend a considerable amount of time on the unpleasant work of correcting and grading homework, tests, and exams \cite{thor10}. While answers to multiple-choice questions do not need major time commitment to correct, free text answers do. With improvements in natural language processing (NLP) and digitalization of educational institutions, the necessary data is available to increase the automation of marking answers. %Not at least due to the COVID-19 pandemic exams are done with computational devices rather than with pen and paper. 

An ``autograder'' system might support teachers in multiple ways. First of all, it might grade questions completely autonomously. Even a seemingly small workload reduction of a few percent in grading effort amounts to thousands of saved working hours per year on a global scale.\footnote{Even a small country such as Switzerland employs more than 120000 teachers (\url{https://en.wikipedia.org/wiki/Education_in_Switzerland}), each spending several hours a year on average with grading.}  Teachers could devote this time to supporting students, e.g., assisting those in need of more personal coaching.  Second, an autograder can be employed when teachers cannot provide feedback to students’ answers due to time constraints. For example, while students are studying at home, answering questions for exam preparation, an autograder might provide an assessment of their answers. %This makes learning more interactive for students.

These are just some of the needs an autograder can serve. However, autograding is only applicable if it achieves a certain level of performance, i.e., its decisions are sufficiently accurate. Otherwise, trust issues arise (known as algorithm aversion) \cite{dzi03,diet15,hsu21,aza20}. Aside from trust, errors by an automatic grading system can have a profound impact on an individual. The student might not only fail the exam, but the entire course and in extreme cases might even have to repeat an entire year of school, meaning separation from existing classmates and psychological distress \cite{cor06}. 
%Misgrading a single question can make all the difference (see page 16 in \cite{rho03}).
Current works, e.g., \cite{vitt20,hsu21} (and also this study), show that about 85-90\% of answers are graded correctly as correct/incorrect if all answers are autograded. Humans make errors as well, and the risk is higher for  repetitive tasks like grading the same question dozens of times. For such tasks, error rates have been reported to be in the range of 0.5\% to 6\% (p. 412 in \cite{smi17}), which is likely lower for exams, since they are often double checked. Thus, 90\% accuracy on graded answers is clearly below human performance and as such not sufficient for autonomous grading. The fact that tuning ``accuracy'' on its own through model improvements is not enough has also been emphasized by \cite{mad18} calling for a broader view on autograding.
% more interdisciplinary view on the problem going beyond technical NLP experts
While errors cannot be fully avoided (at least for now), the type of error should ideally be controllable. For example, it might be preferable to let a student pass who should have failed than to fail a student who should have passed. One reason for this is that given the choice between detecting errors correctly(precision) and detecting all errors(recall), precision is preferable since it leads to a better learning effect for students\cite{nag10}.

%This reasoning is part of legislation in Western countries advocating that a criminal should go free rather than punishing an innocent person. % While we do not proclaim any guidelines, we create transparency and also enable teacher to control the likelihood of each type of error.

Furthermore, from a practical perspective, existing work on autograding appears fragmented and difficult to deploy at a school/university level covering many disciplines. Most papers focus on a single domain (area) and language. Deploying a large number of algorithms (for all kinds of questions and topics) would lead to a system complexity and an IT landscape that is difficult to manage. In the past, less complex, simpler solutions have been preferred to more complex but better-performing solutions \cite{Net12}. This posits another motivation to evaluate existing generic approaches on a larger, broader dataset.

To summarize, reliable autograder systems are necessary for trust and practical utility. Reliability seems difficult to achieve due to the statistical nature of AI and considering that an autograder should work across multiple languages, subjects, and styles of questions and answers. Questions might vary strongly depending on subjects, i.e., a math-related question requires different skills to answer (and to mark) than a question related to other subjects such as language and history. Also, teachers might pose questions very differently, both in terms of syntax and semantic. From a technical perspective, this increases the risk of a mismatch of the training data and (test) data used during operation, i.e., a teacher might pose questions on a novel topic in a way not found in the training data. This can lead to unexpectedly low performance. 

\begin{figure}[htbp]
\includegraphics[width=\linewidth]{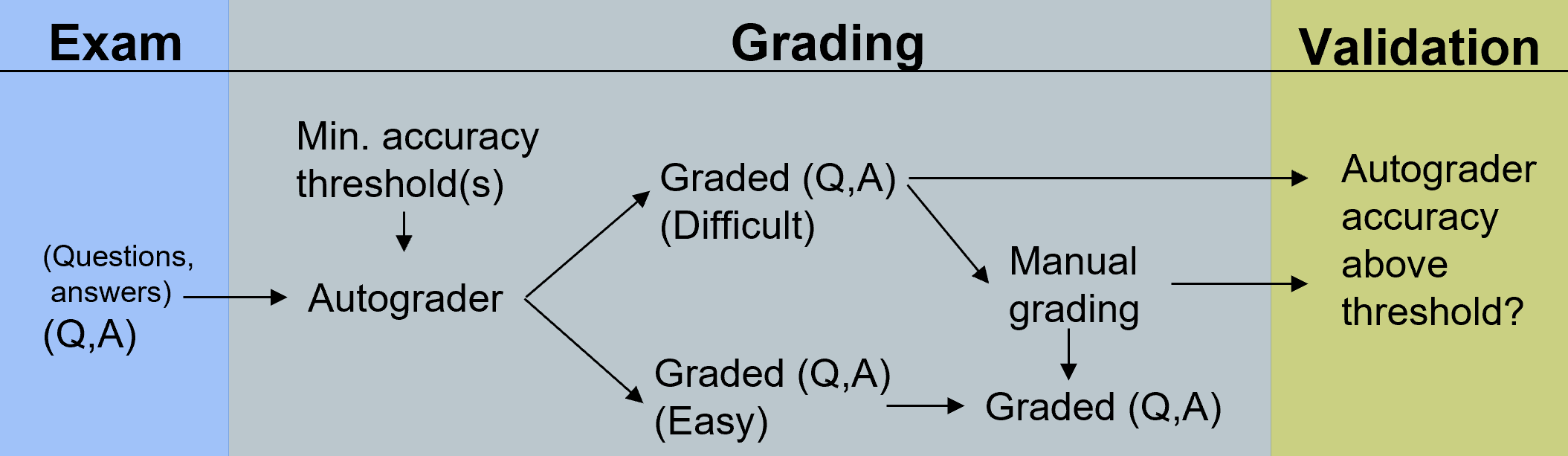}
\caption{Overview of approach: Answers are graded, taking into account minimum accuracy thresholds specified by a teacher. Difficult question/answer pairs are (also) manually graded to ensure that accuracy requirements are met. Accuracy on manually and automatically graded answers is compared to validate the performance of the autograder.} \label{fig:over}
\end{figure}

In this study, we address these questions by using a dataset of question-answer pairs that are more diverse in terms of languages and type of questions, as well as being significantly larger than prior works. We assess multiple state-of-the-art transformer models, namely multilingual BERT and LaBSE, for the task of automatic grading. We compare performance of both models with respect to different types of questions and languages. We restrict ourselves to questions whose answers can be marked as correct or incorrect. We use a similarity score rather than “correct” or “incorrect” as output quantifying the distance of the question-answer pair to the correct (question-)answer pair. Using two thresholds on the score, we can place a question-answer pair into one of three categories: (most likely) “correct”, “uncertain”, i.e., difficult to grade, or (most likely) “incorrect”.  We defer ``uncertain'' question-answer pairs to be graded manually by the teacher. This makes it possible to trade-off the fraction of autograded samples, i.e., work being done (or outsourced) to the autograder, and accuracy metrics on autograded samples. While our results indicate remarkable performance on average for small datasets, i.e., classes with few students and questions, the deviation from the expected performance can be large, i.e., it can be both significantly better or worse. Since a much poorer than expected grading is hardly tolerable, we propose a validation procedure that reduces this risk. It relies on comparing the grading accuracy of a set of questions and answers that are manually and automatically graded. To minimize the effort for teachers, we propose to leveraging the grading information from ``difficult, uncertain samples'' that must, either way, be handled by a human to achieve high accuracy. This comes with a number of concerns we shall elaborate on. Our approach is summarized in Figure \ref{fig:over}.

\noindent In summary, we contribute as follows:
\begin{itemize}
    \item We evaluate existing multilingual transformer models in the context of autograding using a dataset covering multiple languages and question types posed in a different syntax. Our tuned model achieves an accuracy of about 86.5\% when classifying all samples as either correct or incorrect. This is comparable to recent works with a much narrower scope both in terms of questions and languages\cite{vitt20,hsu21}.% as discussed in the related work.
    \item We propose deferring decisions for difficult answers to humans. When doing so, human-like accuracy levels are possible on the remaining answers. 
    \item We also advocate explicitly monitoring and specifying desiderata, on the type of errors an autograder should make. \\
    In combination with deferring answers to humans, this leads to the following outcome: Say we aim for a 2\% false positive error rate (grading an incorrect answer as correct) and 0\% false negative rate –- in other words, our target is to ensure that the autograder does not fail any student unjustly. Then, still about 10\% of all questions are graded. %Asking only for an overall accuracy of 90\% about 45\% of questions are graded.
    
    \item To ensure that autograder performance is not just fulfilled on average on a large set of questions and answers stemming from many exams, but also for a (small) set stemming from a single exam we suggest introducing a validation mechanism. It utilizes existing grading information of difficult questions that were delegated to a human grader. Our experimental evaluation shows that autograder performance on small datasets fluctuates more, i.e., it is indeed commonly lower or larger than on a large dataset. But strong deviations from expected accuracy can be reliably detected.
\end{itemize}

\section{Related Work}

\subsection{Transformers and BERT}
\begin{figure}[htbp]
\centering
\vspace{-12pt}
\includegraphics[scale=.64]{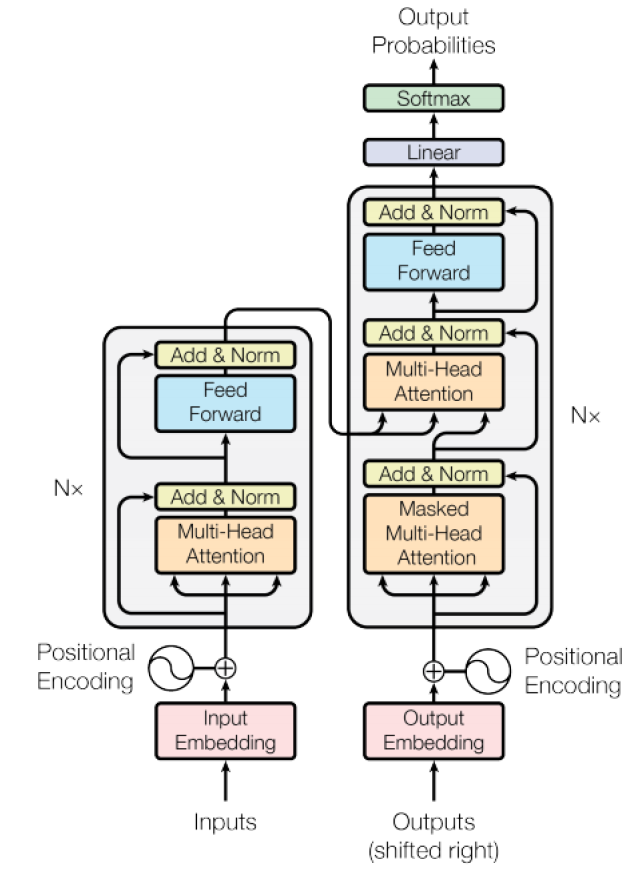}
\vspace{-10pt}
\caption{Transformer (Figure from \cite{Vaswani2017})}
\label{figTrans}
\vspace{-12pt}
\end{figure}

While many techniques from NLP are helpful in autograding, we primarily discuss those using deep learning and, more specifically, transformer models including BERT. Transformers\cite{Vaswani2017} are a key driver for current NLP success. They have a better performance than recurrent and convolutional models for translation problems. Extensions built on transformers such as BERT \cite{lan19} have also yielded remarkable results on question-answering, e.g., the SQuAD benchmark, which makes them appealing for auto-grading. However, the SQuAD (and other benchmarks) are often fairly specific. For SQuAD an answer to a question has to be identified within a given text segment, i.e., the benchmark is a reading comprehension task. It is unclear whether the model works for other types of questions and answers which makes an explicit evaluation necessary. A comparison of transformers and LSTM models suggested that LSTM models might be preferable on small datasets, but, generally, transformers tend to outperform recurrent models \cite{ezen20}. Transformers rely on a self-attention mechanism, as illustrated in the architecture in Figure \ref{figTrans}. %The architecture could be used for translation, e.g., from German (input) to Italian (output). An input could be a single word (in German) and the output a single word (in Italian), which originates from the target sentence shifted to the right. Thus, the provided output word is not the current word to be predicted by the transformer. The output probabilities provide one number for each Italian word, i.e., all words in a dictionary. 

The left part in Figure \ref{figTrans} processing inputs is called an encoder in the context of transformers and the right part of the figure processing (outputs and the encoded inputs) is called a decoder. Attention facilitates remembering ``which past inputs'' are important using weights, rather than ``what information of the inputs'' is important.  While the wording might suggest that this is a subtle difference, in practice, performance gains by remembering ``the address of an object'' rather than ``the object itself'' is large. Technically, a weight is estimated for each past input (based on the current input), indicating its relevance. Past inputs are aggregated using a weighted sum. Transformers also do not require any recurrent computation. While recurrent networks implicitly have positional information (they get one input token after the other and could learn to count them), transformers require an explicit positional encoding with each input. Transformers also leverage embeddings to change ``one-hot'' word-encodings into vectors. 

BERT is a language representation model developed by the Google AI language team \cite{Devlin2018} based on transformers. Technically, BERT is a multi-layer bidirectional transformer encoder for NLP tasks. It performs a bidirectional training of the transformer instead of being bound to right-to-left or left-to-right training. Training relies on the masked language model and next sentence prediction.  15\% of all words are masked and predicted for the masked language model. For the next sentence prediction, the network should learn to tell if for a sentence pair $(A,B)$, sentence $B$ follows directly after sentence $A$ or not. BERT can be trained on any large collection of documents, e.g., Wikipedia. The resulting BERT model can be fine-tuned for numerous NLP tasks such as question-answering. In this case, a pre-trained model is adjusted using question-answer pairs rather than training with sentence pairs.

There also exist multilingual versions of BERT \cite{Devlin2019a} trained on more than 100 languages from Wikipedia articles.  \cite{Reimers2020a} argue that with the multilingual BERT, the ``vector spaces between languages are not aligned, i.e., the sentences with the same content in different languages would be mapped to different locations in the vector space''. This is one motivation to introduce sentence-BERT \cite{Reimers2020a}, which creates sentence embeddings that are close to each other if the original sentences are (semantically) similar.

%Devlin et al. (2019, p. 4173) introduce two variations in size, BERTBASE with 12 layers, 768 hidden layer size, and 12 self-attention heads leading to 110 million trainable parameters and BERTLARGE with 24 layers, hidden layer size of 1024 and 16 self-attention heads leading to 340 million trainable parameters. The pre-training is based on 800 million words from the “BooksCorpus” (Zhu et al., 2015) and 2’500 million words from English Wikipedia articles (p. 4175). BERT uses the embeddings “Word-Piece” as introduced by Wu et al. (2016) with a vocabulary size of 30’000 English language tokens. \\
 %Training is non-trivial, since  not every language has the same number of Wikipedia articles. To address challenges related to an imbalanced dataset, they performed “exponentially smoothed weighting of the data during pre-training data creation” (Devlin, 2019, para. 20) \cite{Devlin2019a}. As an example they state that if 21\% of all data was in English, they would exponentiate it by a factor of 0.7, normalize it again sample from it. This leads to under sampling languages like English and under sampling smaller languages. “In the original distribution English would be sampled 1000x more than Icelandic, but after smoothing it’s only sampled 100x more” (Devlin et al., para. 20) \cite{Devlin 2019b}. 

%Tokenization wise a WordPiece based vocabulary with 110’000 tokens is used. There are no labels that indicate the language to allow for zero-shot training. $F^i$nally, all words/ letters in the vocabulary are lower cased and all accent characters removed.

Besides the multilingual version of the classic BERT, there is also a Language-agnostic BERT Sentence Embedding (LaBSE) \cite{Yang2020}. It is a multilingual version of BERT with sentence embeddings similar to Sentence-BERT\cite{Reimers2020b}.  The LaBSE model claims to align vectors from different languages that are semantically similar. %It uses the same hyperparameters and architecture as the multilingual BERT base stated above. However, the tokenizer and data that the model was trained on differ. The LaBSE tokenizer has 501’153 tokens.
LaBSE is based on a dual encoder architecture. While the two input sentences are individually encoded, they use a shared pre-trained BERT encoder. It was trained on mono- and bilingual translation pairs from Wikipedia and CommonCrawl, i.e., 17 billion monolingual sentences. The bilingual translation pairs are sourced from various web pages. %Moreover, \cite{Reimers2020a} set the maximum number of sentence pairs for each language to 100 million. There are 109 different languages with unequal distribution of representations in the corpus. 

\subsection{Autograding}
Automatic grading involves multiple technologies and variations of tasks. We focus on Automatic Short Answer Grading (ASAG)\cite{bur15}, but there is also research on automatic essay scoring (AES) \cite{dik10}. As the name implies, ASAG handles less data than AES, i.e., typically only one phrase to one paragraph, but more than ``fill-the-gap'' (FG) type questions, which consider only one to few words\cite{bur15}. Moreover, the focus lies on content rather than writing style and structure (AES) or single words (FG). ASAG (though potentially also AES) includes computer science \cite{Sultan2016} and mathematical questions, including math-related questions with open-ended answers\cite{bara21}. It also encompasses questions from other areas such as psychology and biology \cite{Riordan2018,att08}.

AES has a long history, dating back to 1964. Back then, 30 variables were identified and included in an early system for AES\cite{Page1967}. In short, the ``PEG'' system ``scores based on shallow linguistic features, including essay length, number of prepositions, number of relational pronouns, variation of word length, etc.'' using a simple KNN (K-nearest neighbors) approach \cite{Bin2008}. A system called IEA \cite{Landauer2013} uses Latent Semantic Analysis (LSA) to analyze and score essays in various languages as well as providing feedback to exercises. Other approaches like the “e-rater”\cite{Burstein2014} or \cite{bas13} have multiple features, e.g., to tag words, identify discourse (based on cue-words) and perform topical analysis. In particular, \cite{bas13} learns a similarity metric between answers. In contrast, we rely on an embedding of text and standard similarity metrics, i.e., the cosine similarity. However, we might also learn a similarity metric using additional features stemming from early and more recent works such as \cite{bur15}. \cite{bur15} employs clustering of answers so that a human only has to grade one answer of the cluster, which saves a significant portion of time. This idea might also be added to our approach.  
A neural network-based approach to fine-tune a BERT model is discussed in \cite{Yang2020a,wang21}. \cite{Mayfield2020} states that pre-trained models may generalize better on new essay topics that the trained model has not seen yet than simple neural networks such as LSTMs.% However,  \cite{Mayfield2020} also put the usefulness of BERT for AES into question “Fine-tuning BERT produces similar performance to classical models at significant additional [computational] cost”.  

%Among others, the “c-rater” (c stands for content) is such a grader. It also has a modular structure like the ‘e-rate’. \cite{Leacock2003}.%  based on the Earth Mover Distance (EMD). EMD is similar to the “Wasserstein distance” and “Mallows” \cite{Levina2001}

%\cite{Kumar2017} uses a siamese bidirectional LSTM together with a pooling layer.
%  instead of training it on a more general data set or not at all

For ASAG, \cite{Shehab2018} compares different automatic grading algorithms used for Arabic free text answer questions. They identify string-based and corpus-based text similarity approaches. \cite{Sultan2016} utilizes recent measures of lexical similarity and aggregation of word embeddings \cite{Baroni2014}  to outperform methods like LSA for short answer grading. \cite{Suzen2020} developed a model that predicts ``marks using the distance between the model answer and the student answer''. \cite{Suzen2020} assumes that the marks given to answers highly depend on the similarity of words between the student answers and the teachers' solutions. The paper uses k-means clustering for sorting answers into similar grade groups following the idea of \cite{bur15}. \cite{Riordan2018} compares neural network based approaches being used for essay scoring with short answer scoring. The paper argues that LSTM based neural networks perform mostly better than non-neural network based models for short answer scoring. \cite{Sung2020} base their research on a pre-trained BERT model instead of simpler neural networks, i.e., LSTMs as in \cite{Riordan2018, Kumar2017}. It is argued that the pre-trained BERT performs better when data augmentation is done with domain-specific data. For instance, to grade short answers on the domain of psychology, training BERT on a data set with domain data from psychology will result in better performance for grading psychology-related short answers. These findings also highlight that BERT might not perform as well on a very diverse dataset (like ours) as on a very domain-specific one. \cite{bara21} also obtained best results (out of a set of models) using BERT for open-ended mathematical questions. 

\cite{vitt20} focused on a small-scale study using AI for formative and summative assessment for a data science class. They employ both static code analysis for code written in the data science language ``R'' and embeddings using BERT based on students' responses and correct answers. When using embeddings only, they report an accuracy comparable to ours, i.e., 87\% for balanced data. As in our work (and others), similarity of embeddings was measured using the cosine distance. In contrast to our work, the dataset is only small, monolingual, and from a particular domain only. \cite{vitt20} also assesses the value of feedback provided by the autograder based on student responses to a questionnaire and learning outcomes. The paper reports a reduction in grading time when a teacher was able to use autograding results. However, they did not check, whether the graded answers are indeed correct. Thus, it is unclear whether time savings came at the cost of increased errors by relying too much on the autograding answers. We aim at a system design that grades some questions automatically without human intervention or double-checking by achieving (a large) user-specified precision and avoiding automatic decision-making in case of uncertainty. 

Trusting systems based on artificial intelligence (AI) or, more specifically, NLP, is a widespread concern beyond autograding. AI is known for its ``black-box'' nature, making it difficult to understand. This led to many works related to explaining such networks and trust, e.g., \cite{sch19, sch18,hsu21}. 
\cite{hsu21} used a simple logistic regression model as autograder. The paper investigated student perceptions at college level of an autograder achieving 90\% accuracy for questions related to programming (Python code). It showed that students overestimated the probability of the autograder misjudging correct examples as wrong (False Negative (FN)), which is a problem with respect to trusting the system. However, despite tuning to achieve low FN rate, the work \cite{hsu21} could not address this problem, i.e., the FN rate was still 10\%. Incorrectly failing 10\% is undoubtedly too high since even lower errors rates have led to lawsuits and outrage (see \cite{rho03}, p. 16). Our work addresses this problem. We allow setting a FN rate of even 0\%. We maximize the number of graded answers, while ensuring that this constraint is fulfilled. In a programming class, \cite{aza20} investigated strategies to deploy an unreliable autograder during actual exams. Their approach was to grant students multiple attempts, which lowered the FN rate.

Explainability techniques have also been assessed in the context of autograders \cite{kum20exp}. Explanations can serve as justifications for (or, at least increase understanding) of automatic grading decisions.  \cite{miz19} used word embeddings and an LSTM with an attention mechanism to also focus on justification (for a response). There are also attempts to address short-comings of AI with respect to fairness\cite{frie19}. In particular, for auto-grading \cite{mad17} discussed fairness, looking into bias as well as test design and score interpretation.

% a lack of transparency and unethical decision making with regulation, e.g., the General data protection regulation grants individuals the ``right to explanation'' for automated decision-making, or technical means, e.g.
In summary, the above approaches touch upon different aspects of an autograder system. Features relevant for autograding can be human-defined linguistic features, features from decompositions such as LSA, and features learnt in neural network architectures. While many types of classification algorithms have been used for autograding, deep learning based models are prevalent. 
Simple string matching-based approaches and similarity metrics based on embeddings of answers using neural language models have been used to assess the match between a given answer and a ground truth answer.
Furthermore, several papers expressed the idea of clustering, e.g., to grade similar answers to the same question jointly. \\
To conclude, BERT has been employed for both ASAG and AES. Assessing multilingual models and large datasets from various subjects and a rich set of instructors is a new development. In contrast to \cite{hsu21,aza20,vitt20} our design follows the philosophy that trustworthy results are more important than grading all questions.
Furthermore, most works did not consider ethical concerns. They did not investigate the option of not grading answers to achieve high accuracy for actual predictions, which is key to ensuring trust in autograders. 

\section{Data}
\begin{figure}[htbp]
\includegraphics[width=\linewidth]{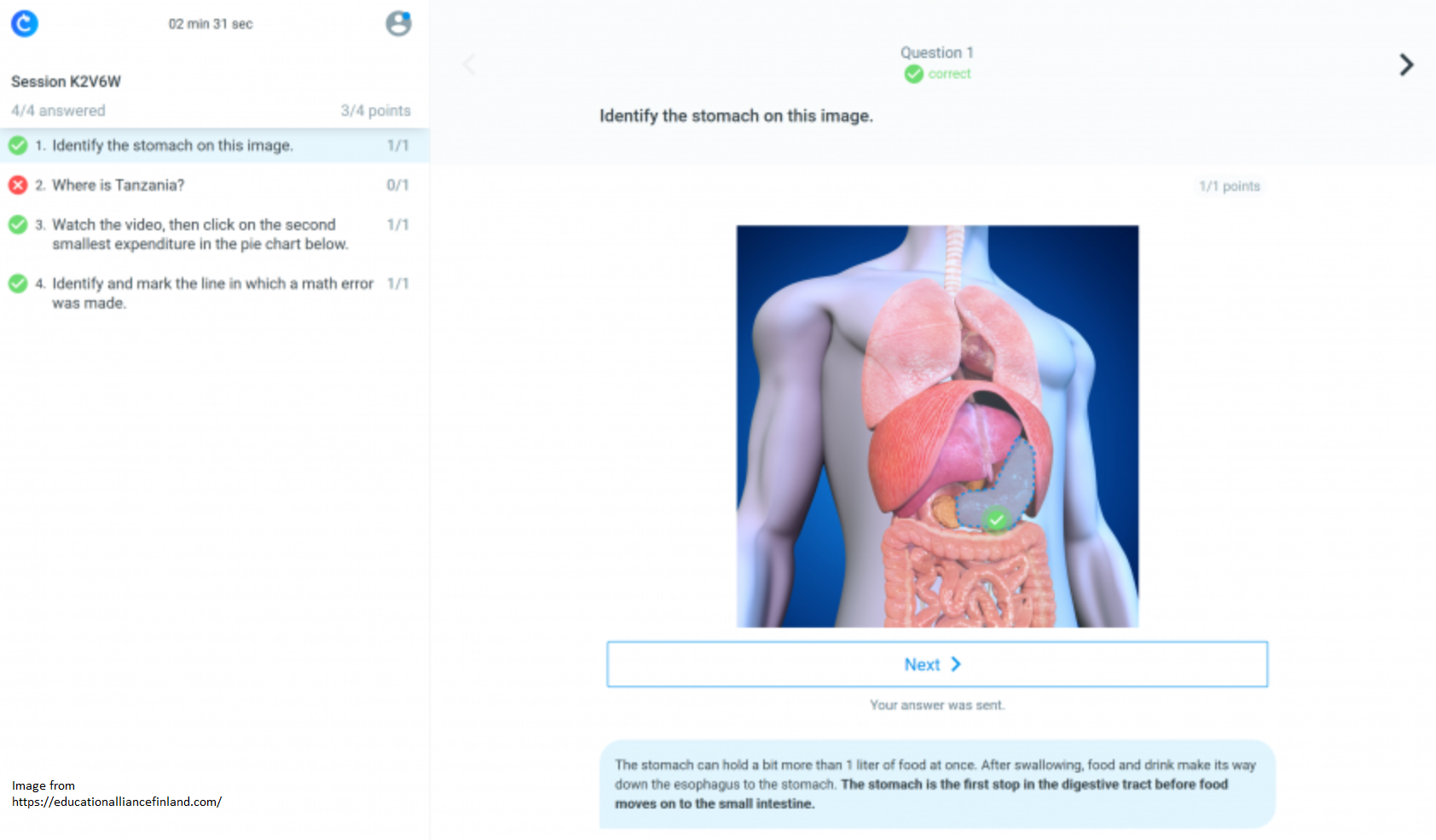}
\caption{Web interface of tool for data collection from Classtime. It allows for many types of questions. In this study, we only use questions allowing for free text answers.}
\label{fig:clas}
\end{figure}

The data set stems from the startup Classtime\footnote{\url{www.classtime.com}}. Classtime provides a web browser-based assessment tool illustrated in Figure \ref{fig:clas}. Its features include a free text answer question type, which is the focus of this study. The tool allows students to write answers ranging from one to hundreds of words or numbers. Teachers can compose their own questions and answers or select them from a library of existing questions. 

\begin{figure}[htbp]
\includegraphics[width=0.6\linewidth]{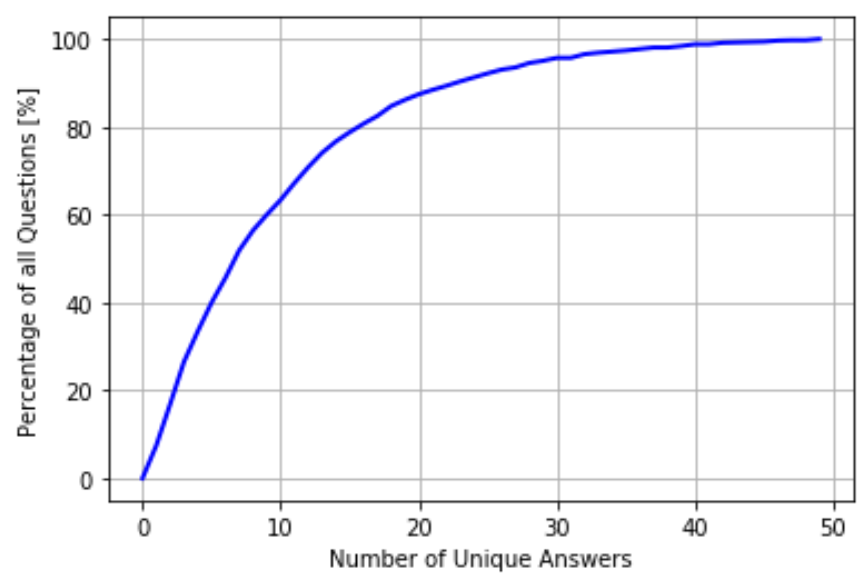}
\caption{Cumulative distribution of the number of unique answers per question  } %A ``token'' can be either a number or a word.
\label{fig3c}
\end{figure}

In this study, we restrict ourselves to questions requiring students to enter textual (or numerical) answers. The raw data set consists of a set of tuples $(Q,A^c,A^g,G)$, where $Q$ is a question, $A^c$ the correct answer by the teacher, $A^g$ the answer to grade (provided by a student) and $G \in \{correct,incorrect\}$ a mark by a teacher for the student's answer $A^g$. The dataset consists of approximately 10 million answers and 990000 questions. Each question typically has multiple answers as indicated by the cumulative distribution of the number of questions based on unique answers in Figure \ref{fig3c}. The distribution highlights that about 17\% of all questions have two or fewer different answers. More than 50\% have seven or more unique answers. 

The answer and grade provided by the teacher are taken as ground truth. Thus, our system heavily relies on the quality of grading by teachers. To assess data quality, we manually investigated 300 tuples. We did not find errors in the teachers' answers and few in their grading, as we will discuss more in the evaluation. But we found several decisions where the grading could be deemed lenient or hard at first sight. However, we lack the necessary context, e.g., rubrics and background on students. For example, in many countries students are separated based on skills leading to different expectations of answer quality for identical questions.

\begin{figure}[htbp]
\centering
\includegraphics[scale=0.7]{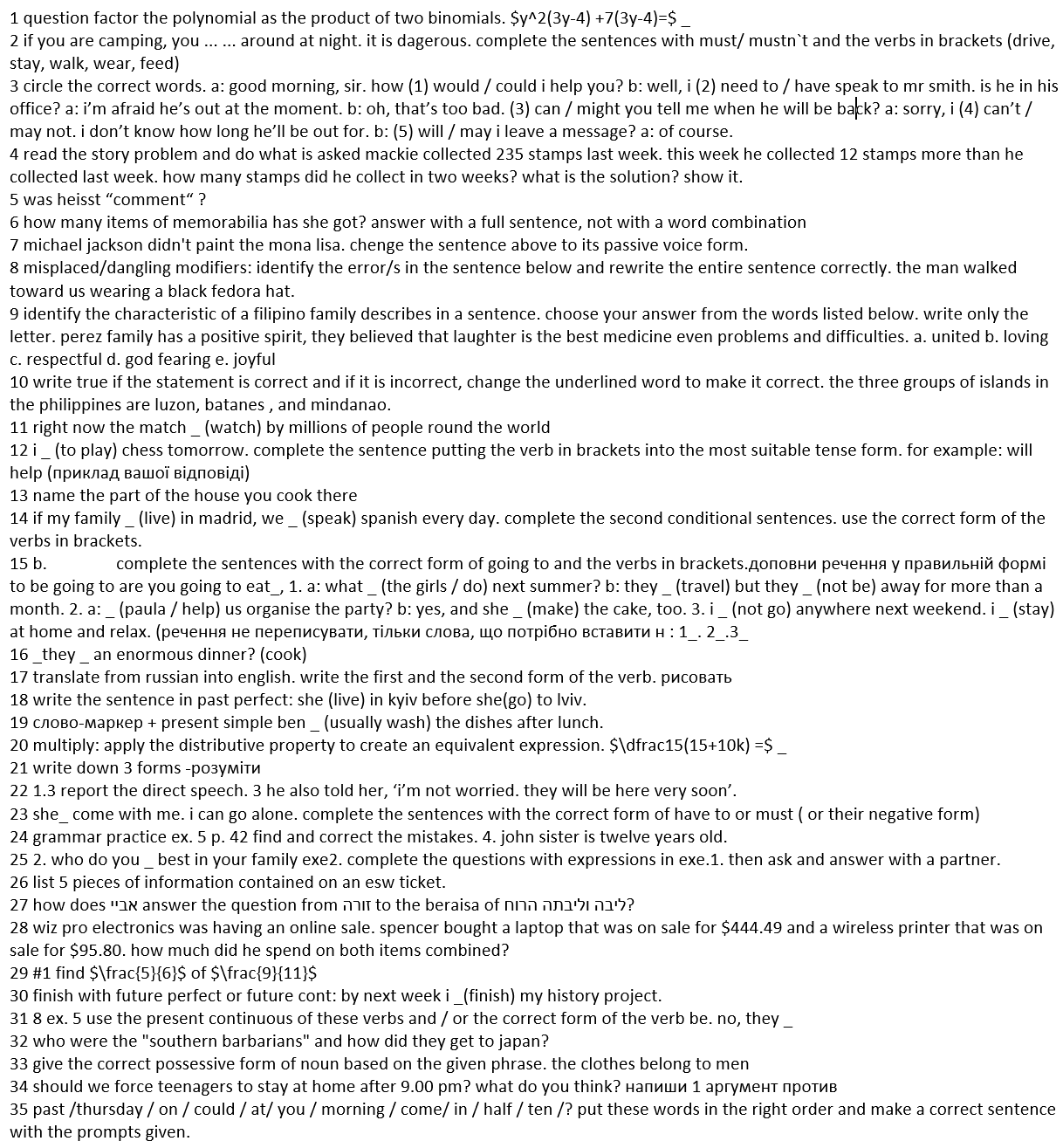}
\caption{English question samples in lowercase.} % Questions might also have a title, which is not shown.
\label{fig2a}
\end{figure}

Two exemplary questions with multiple answers are shown in Table \ref{tab1}. The first question could be marked correctly by comparing the ground truth and the student's reply. However, the second question cannot be marked correctly using a simple binary comparison but requires, for example, the notion of similarity. We also provide additional sample questions in Figure \ref{fig2a}. Not unexpectedly, they highlight the diverse nature of questions in content but also in style and syntax. For example, though we only show questions in English, they might still contain elements of other languages. Some math questions contain special syntax or markup code, e.g. ``LaTex'', others do not. For language questions, where the correct form of a verb should be inserted (fill-in-the-gap), a verb might be given directly after the gap or after the question or not all. Instructions might be part of the question text or missing, i.e., assumed to be clear. The lack of a common uniform structure might be seen as a data quality issue. It might also hinder the application of specific algorithms for specific question types if they rely on a pre-defined question and response structure. The chosen questions in Figure \ref{fig2a} are posed in English. Language samples constitute roughly 50\% in Figure \ref{fig2a}. They are overrepresented since English is the number one foreign language worldwide and exam questions tend to be posed in English even for non-native speakers.\footnote{We also translated questions from other languages to English. We found significantly fewer questions related to language (though they are still common).}

\strutlongstacks{T}
\begin{table}[htbp]
%\begin{adjustbox}{max width=\textwidth}{%

\scriptsize
\begin{tabular}{|l|l|l|l|l|}

\hline
\textbf{\Longstack{Simil-\\arity}} & \textbf{Question}                                  & \textbf{Ground Truth}                                       & \textbf{Student Reply}                                                & \textbf{\Longstack{Corr-\\ect?}} \\  \hline

0.788               & \Longstack{i can't \\ stand \_ (sit) \\ with carla, she \\ is such ...} & sitting                                                     & sitting                                                               & Yes                          \\ \hline
0.507               & \Longstack{i can't \\ stand \_ (sit)\\ with carla, she \\ is such ...} & sitting                                                     & to sit                                                                & No                          \\ \hline
0.455               & \Longstack{i can't \\ stand \_ (sit)\\ with carla, she \\ is such ...} & sitting                                                     & set                                                                   & No                          \\ \hline
0.997               & \Longstack{define the \\ following: \\business manager}             & \Longstack{handles finances \\ and accordingly \\ as a part of \\ a artist team} &  \Longstack{member of  the \\ artist's team that \\ handles the finances \\ and accounting.} & Yes                          \\ \hline
0.997               & \Longstack{define the \\ following:\\ business manager}             & \Longstack{handles finances \\ and accordingly \\ as a part of \\ a artist team} &  \Longstack{handles \\ finances/\\ accounting for \\ the show}                              & Yes                          \\ \hline
0.992               & \Longstack{define the \\ following:\\ business manager}             & \Longstack{handles finances \\ and accordingly \\ as a part of \\ a artist team} &  \Longstack{handles money \\ and accounting on \\ artists team}                          & Yes                          \\ \hline
\end{tabular}
%\end{adjustbox}
\caption{Example questions and answers. Each row corresponds to a training sample.} \label{tab1}
\end{table}

\begin{table}[htbp]

\center
\begin{tabular}{|l|l||l|l|}
\hline
\textbf{Language}  & \textbf{\Longstack{Question\\count}} & \textbf{Language}  & \textbf{\Longstack{Question\\count}} \\ \hline
Ukrainian                                     & 523257         &Spanish                                       & 4421           \\ \hline
English                                       & 184069         &Croatian                                      & 3557           \\ \hline
Russian                                       & 179177         &English US                                 & 1744           \\ \hline
German                                        & 71282          &Thai                                          & 1073           \\ \hline
Estonian                                      & 10028          &Czech                                         & 647            \\ \hline
French                                        & 5799           &Italian                                       & 624            \\ \hline
Portuguese                                    & 5605           &Turkish                                       & 447            \\ \hline

\end{tabular}
\caption{Overview of question counts per language.}\label{tab2}
\end{table}

% Figure~\ref{fig1} illustrates that most questions are fairly short.

% \begin{figure}[htbp]
% \includegraphics[scale=0.6]{question_word_distro.png}
% \caption{Words occurring in question (and its title}
% \label{fig1}
% \end{figure}
\begin{figure}[htbp]
\centering
\includegraphics[scale=0.6]{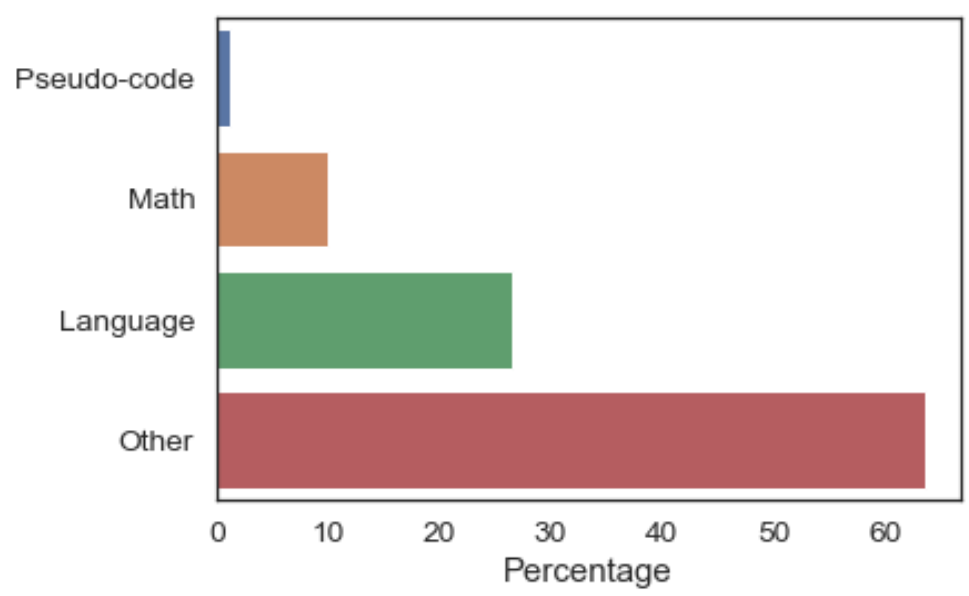}%three_types_all_data.png}
\caption{Distribution of question types for $n=955$ manually classified samples.}
\label{fig3a}
\end{figure}

\begin{figure}[htbp]
\centering
\includegraphics[width=0.7\linewidth]{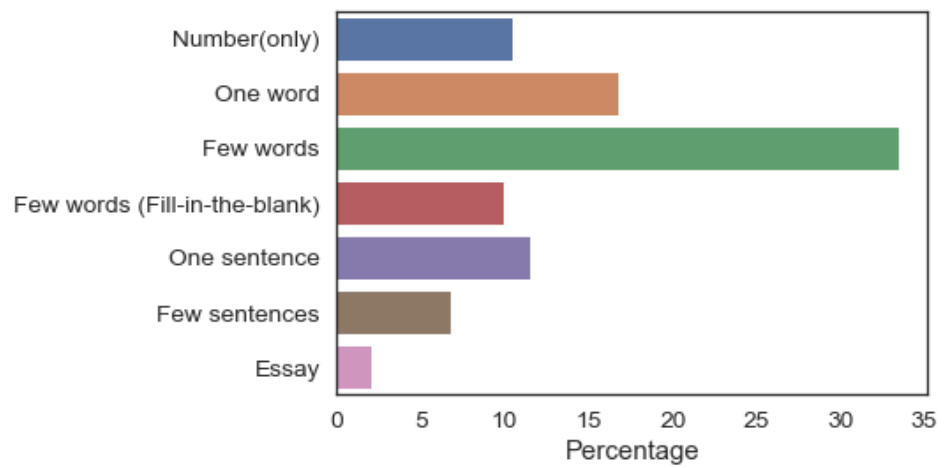}%three_types_all_data.png}
\caption{Distribution of questions by answer format for $n=955$ manually classified samples. } %A ``token'' can be either a number or a word.
\label{fig3}
\end{figure}

\begin{figure}[htbp]
\centering
\includegraphics[scale=.7]{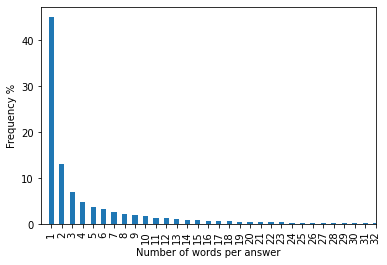}
\caption{Distribution of word counts of answers (limited to 32)}
\label{fig2}
\end{figure}

The data stems from multiple languages, Ukrainian being the most represented with more than 500'000 questions (see Table~\ref{tab2}). The distribution of question types is stated in Figure \ref{fig3a}. A significant portion of questions is related to language including both questions related to foreign languages and the native language of a student. There is also a large number of math-related questions but very few coding questions. Other questions are a mix of a large pool of subjects, e.g., biology, art, physics, economics, law, geography, chemistry, history, and sociology.

%We classified based on the content of the answer into answers that only consist of numbers, only of text, i.e., anything but numbers, and a mix of both. A manual investigation of the data yielded that there is a significant portion of math-related questions, some of which only demand a single number as reply.

The distribution of answers based on response format is shown in Figure \ref{fig3}. Fill-in-the-blank type questions are mainly related to language. An answer of a few words is not necessarily shorter than an answer of one sentence but rather the grammatical structure was not that of a sentence. A concise picture of the distribution of the word count of answers is given in Figure \ref{fig2}. The frequency of answers is quickly decaying with word count, i.e., most answers are only a few words in length. %Another categorization based on numerical, textual, or mixed answers is illustrated in Figure \ref{fig3b} in the Appendix.

Many questions answer pairs are duplicates, i.e., different students often provided the same answers. We removed duplicates in the dataset resulting in 4.3 million question-answer pairs.  %They might be ``true'', ``false'', ``yes'', or ``no'', though more commonly, short answer indicate 

%The distribution of the word count of answers is stated in Figure \ref{fig2}. Many answers are fairly short. Around 58\% of the answers are correct.%They might be ``true'', ``false'', ``yes'', or ``no'', though more commonly, short answer indicate 

In our dataset, there are more correct than incorrect answers, i.e., around 58\% of all answers are correct. In machine learning, using imbalanced datasets can have adverse effects on performance.  Balancing refers to creating an equal distribution of the classes, i.e., 50\% correct and 50\% incorrect answers. We employ oversampling to create the same number of correct and incorrect replies. We randomly pick answers from other questions. Picking a random answer for a specific question out of millions of answers likely results in the answer being incorrect for the question.

\section{Model and Definitions} \label{sec:defs}
\begin{figure}[htbp]
\centering
\includegraphics[scale=0.3]{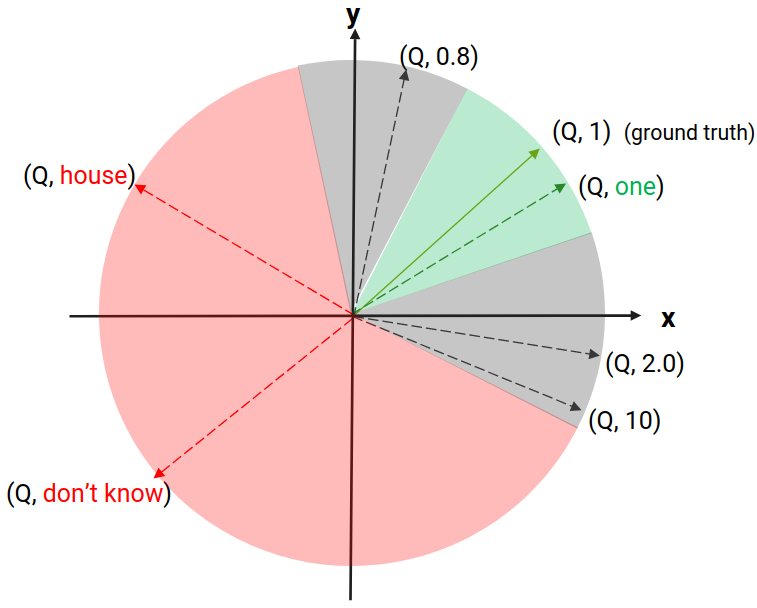}
\caption{Vectors for question-answer pairs (Q,A) with Q being “Calculate 0.2+0.3+0.5+1-1.2+0.2”. Vectors falling in the green area are classified as correct, those in the red area as incorrect and those in the grey area deemed ``uncertain''. }
\label{fig4}
\end{figure}

We use the multilingual BERT and LaBSE models as discussed in the related work. Both models take as input two pairs $p_c=(Q$,$A^c)$ and $p_g=(Q$,$A^g)$, where $Q$ is a question, $A^c$ is the correct answer and $A^g$ the answer to grade. The output are two vectors, i.e., a vector $v_c$ for the pair $p_c$ and a vector $v_g$ for the pair $p_g$. While LaBSE performs sentence embeddings, for BERT we averaged the vectors for all words. The model's objective is to maximize the similarity between vectors $v_c$ and $v_g$ if $A^g$ is correct, while vectors $v_c$ and $v_g$ should be dissimilar otherwise. This is illustrated in Figure~\ref{fig4}.

%\cite{Garbacea2014}

The loss\cite{Barz2020} for two question pairs $p_c, p_g$, ground truth value $y$ and hyper-parameter $margin$ is given by:
\begin{equation*}
\small
loss(p_c,p_g, y) = \left\{ \begin{array}{cl}
1-{cos}(p_c,p_g), &   if \ y = 1 \\
max(0,{cos}(p_c,p_g) - margin), &  \ if  \ y = -1 
\end{array} \right.
\end{equation*}
where ${cos}(a,b)$ is the cosine similarity. %defined as follows:

Similarity $s(Q,A^c,A^g)$ for a question-answer triplet is indicated by the angle between vectors $p_c$ and $p_g$, i.e., $s=p_c\angle p_g$. If the angle is small, the given answer is likely correct. One can find a threshold $T^*$ that maximizes accuracy by grading a question-answer with a similarity above the threshold as correct and incorrect otherwise. Near this threshold, uncertainty of predictions of the autograder is largest, i.e., the autograder might appear to be no better than guessing. In other words, such question-answer pairs are difficult to grade for the autograder. To improve the autograder, we deem these pairs as unsuitable for autograding, i.e., we exclude question-answer pairs falling in a specific range of similarities (around the threshold $T^*$) from being classified. The range is given by a lower and an upper bound [$T^i$, $T^c$]. Formally, for a dataset $D$, we define the subsets of (Q,A) with larger or lower similarity than a threshold $T^*$ and those within a threshold in the range [$T,T'$] as:
\begin{align*}
&D(T,>):=\{(Q,A^c,A^g,G) \in D | s(Q,A^c,A^g)>T\}\\ 
&D(T,<):=\{(Q,A^c,A^g,G) \in D | s(Q,A^c,A^g)<T\}\\
&D(T,T'):=\{(Q,A^c,A^g,G) \in D | (s(Q,A^c,A^g)\geq T) \wedge (s(Q,A^c,A^g) \leq T')\}
\end{align*}

Thus, the dataset $D(T^i,<)$ consists of all question-answer pairs with a similarity $s(Q,A^c,A^g)$ below the lower threshold $T^i$. These question-answer pairs are judged as easy to grade and classified as incorrect. Analogously, samples in $D(T^c,>)$ are graded as correct. 

The accuracy $C_{D,Easy}^c(T^c)$ on samples classified as correct for a threshold $T^c$ is given by: $$C_{D,Easy}^c(T^c):=\sum_{\substack{(Q,A^c,A^g,G) \in D(T^c,>),\\ G='correct'}} \frac{1}{|D|}$$
The accuracy $C_{D,Easy}^i(T^i)$ on samples classified as incorrect for a threshold $T^i$ is given by: $$C_{D,Easy}^i(T^i):=\sum_{\substack{(Q,A^c,A^g,G) \in D(T^i,<),\\ G='incorrect'}} \frac{1}{|D|}$$
We also use the accuracy on difficult samples that are classified as either correct or incorrect using the optimal single threshold $T^*$:

\begin{align*}
C_{D,Diff}^c(T^*,T^c)&:=\sum_{\substack{(Q,A^c,A^g,G) \in D(T^*,T^c),\\ G='correct'}} \frac{1}{|D(T^*,T^c)|}\\
C_{D,Diff}^i(T^i,T^*)&:=\sum_{\substack{(Q,A^c,A^g,G) \in D(T^i,T^*),\\ G='incorrect'}} \frac{1}{|D(T^i,T^*)|}
\end{align*}

\begin{figure}[htbp]
\includegraphics[width=\linewidth]{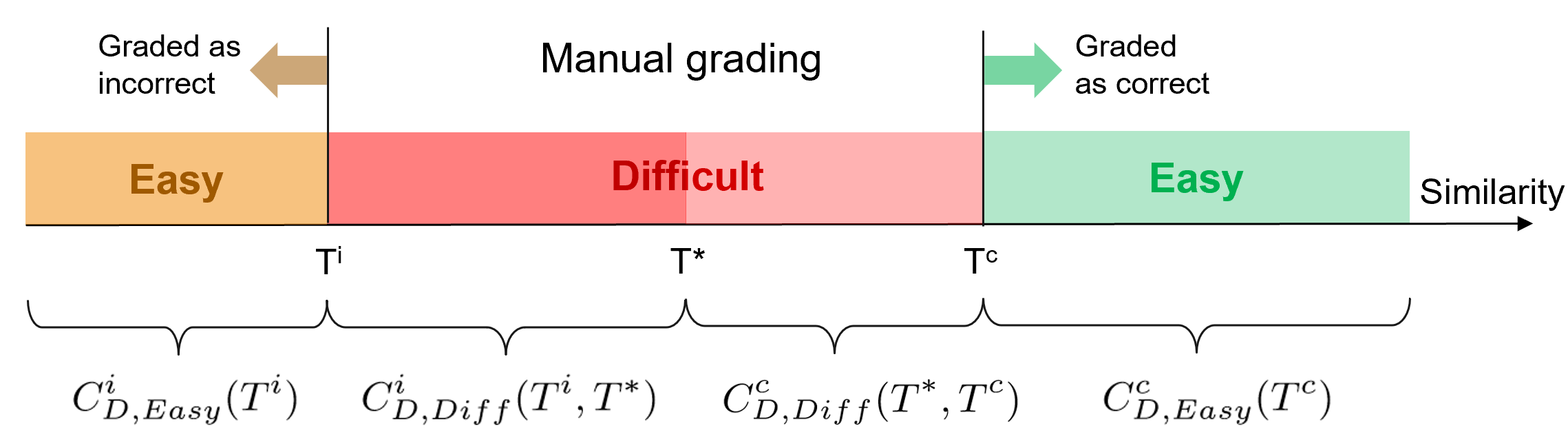}
\caption{Illustration of definitions for data (within a similarity range) used to compute  $C_{D,Diff}^c(T^*,T^c)$,  $C_{D,Diff}^i(T^i,T^*)$,  $C_{D,Easy}^i(T^i)$  and  $C_{D,Easy}^c(T^c)$} \label{fig:sim}
\end{figure}

We illustrate the data used to compute accuracy $C_{D,Diff}^c(T^*,T^c)$, $C_{D,Diff}^i(T^i,T^*)$, $C_{D,Easy}^i(T^i)$ and $C_{D,Easy}^c(T^c)$ for exemplary thresholds $T^c$, $T^i$ and $T^*$ in Figure \ref{fig:sim}. Figure \ref{fig8} in the evaluation shows actual accuracy values $C_{D,Easy}^i(T)$ and  $C_{D,Easy}^c(T)$ for a variable threshold $T$.

A teacher can specify similarity thresholds $T^i$ and $T^c$ to control the type of error being made, e.g., she might specify that a correct answer should never be misclassified as wrong. However, since similarity thresholds are difficult to relate to actual performance, i.e., accuracy, we let a teacher specify a minimum accuracy level $C_{min}^i$ for samples classified as incorrect or, equivalently put, a maximum false-negative rate. Similarly, for samples being classified as correct a teacher can specify a minimum accuracy level $C_{min}^c$, equivalently put, a maximum false positive rate. The accuracy thresholds $C_{min}^i$ and $C_{min}^c$ can be used to compute similarity thresholds $T^i$ and $T^c$ as follows:\\ 

We want to find the maximum fraction of samples that can be automatically graded while fulfilling the constraints on accuracy, i.e., 
\begin{align*}
T^i&:=\argmax_{T} (C_{D,Easy}^i(T)> C_{min}^i)\\
T^c&:=\argmin_{T} (C_{D,Easy}^c(T)> C_{min}^i)
\end{align*}
A very simple optimization procedure uses all similarities $S$ as candidates for $T$ and picks the largest one for $T^i$ and the smallest one for $T^c$. \footnote{Investigating similarities in sorted order combined with efficient search algorithms such as binary search allows for a fast implementation.}

Assume the best achievable threshold $C_{min}^*$, when all data is graded, is obtained for a (tuned) threshold $T^*$. If both accuracies $C_{min}^i$ and $C_{min}^c$ are larger than $C_{min}^*$ then it holds that $T^i<T^*<T^c$, i.e., the samples within range $[T^i,T^c]$ are considered as difficult and deferred to a human. However, in principle, one of the accuracies $C_{min}^i$ and $C_{min}^c$ could be set below $C_{min}^*$. For example, the autograder might grade on average 99\% of all questions correctly. A teacher might accept an accuracy of 98\% for $C_{min}^c$ but want 100\% for $C_{min}^i$. In this case, we might have $T^c<T^*$ and $T^i<T^*$ and also $T^c<T^i<T^*$. Then there are no difficult samples that must be manually graded since the range $[T^i,T^c]$ is empty. Furthermore, we can increase $T^c$ to $T^i$ to improve accuracy $C^c_{D,Easy}$ without an increase in samples to grade manually, i.e., we can set $T^c:=T^i$. %Generally, we can define $T^c:=\max(T^c,T^i)$ and $T^i:=\max(T^i,T^c)$.

We define $F^i$ as the fraction of all samples in $D$ being classified as incorrect, i.e., $F^i:=|D(T^i,<)|/|D|$.  The terms $T^i$, $C^i$ and $F^i$ and $C_{min}^i$ are illustrated in Figure~\ref{fig6}. The case for $F^c$, $T^c$, $C^c$, and $C_{min}^c$ is analogous, in particular, $F^c:=|D(T^c,>)|/|D|$. Thus, the sum of $F^i$ and $F^c$ gives the total fraction of classified samples.

\subsection{Autograder Validation} \label{sec:val}
So far, we have described how to raise the (average) accuracy of an autograder by assigning difficult question-answer pairs to humans. While this is important, it might not be sufficient to inspire complete trust in a system. Even if the autograder's performance has been assessed properly during development, i.e., using an (independent) large test set, additional validation for each usage, i.e., checking the autograder's performance for each exam, is recommended for two reasons: First, performance guarantees are statistical. In particular, for a small dataset, i.e., one with few questions and students, large deviations from average performance cannot be ruled out completely. This can be offputting since  no exam should be very poorly graded. Second, the variation in possible question-answer pairs is large and likely not captured even within a large training and test set. Natural language allows for very diverse type of samples, teachers might have a unique style of asking questions and, additionally, knowledge constantly evolves, i.e., new terms and concepts are continually being invented.
Performance of an autograder may drop considerably for seemingly minor differences in training and test data distribution.%Machine learning is subject to a problem called concept drift, i.e., 

Suppose a teacher should mark a set of question-answer pairs $D_M$. Then, the performance of the autograder on the graded part of $D_M$, i.e., the answers graded as incorrect $D_M(T^i,<)$ and those as correct $D_M(T^c,>)$, might be significantly better or worse than indicated by the teacher's supplied accuracy thresholds $C_{min}^i$ and $C_{min}^c$. We aim to validate that actual performance on the autograded samples to ensure that the two constraints $C_{D_M,Easy}^c(T^c)>C_{min}^c$ and $C_{D_M,Easy}^i(T^i)>C_{min}^i$ are fulfilled. Unfortunately, we do not know the ground truth for any sample in $D_M(T^i,<)$ or $D_M(T^c,>)$ and we also do not like to impose additional work on the teacher to grade any of these questions. From an efficiency perspective, it is preferable to leverage samples that have to be graded by the teacher anyway (if they exist), i.e., those that are deemed difficult to grade.

% A teacher can validate the claimed performance of an autograder by manual grading a sample of all answers, followed by a comparison of the manually graded replies and the automatically graded questions. More precisely, she can randomly sample from both datasets $D_M(T^c,>)$ (answers graded as correct by autograder) and $D_M(T^i,<)$ (answers graded as incorrect) to get datasets $D_R(T^c,>)$ and $D_R(T^i,<)$. She can then compute the autograder's accuary $C^c_R$ and $C^i_R$ on these datasets using her manual grading as ground truth. Finally, she checks if accuracy constraints are met, i.e., $C^c_R> C_{min}^c$ and $C^i_R>C_{min}^i$. Since these tests are performed on a sample, statements are probabilistic, i.e., we get a probability the result is valid on the entire data to grade $D_M$ given that it holds on the sample, i.e., $p(C^c_M> C_{min}^c| C^c_R> C_{min}^c)$. The desired confidence in a statement $C^c_M> C_{min}^c$ can be controlled by the number of samples $|D_R(T^c,>)|$ and $|D_R(T^i,<)|$.

Treating the teachers grading on the difficult samples as ground truth we can estimate the system's accuracy on easy to grade samples. We relate the accuracy on data deemed easy to grade and graded as correct $C_{D_M,Easy}^c(T^c)$ with that on data deemed difficult to grade. More precisely, we focus on ``difficult to grade'' data that would be graded as correct for a single threshold $T^*$, i.e., the data within similarity range [$T^*,T^c$] (see Figure \ref{fig:sim} yielding accuracy $C_{D_M,Diff}^c(T^*,T^c)$. (Analogous thoughts apply for $C_{D_M,Diff}^i(T^i,T^*)$ and $C_{D_M,Easy}^i(T^i)$)

Specifically, we assume (to be verified in the evaluation) that if the accuracy on the difficult to grade samples matches at least the one on difficult samples of the dataset used to test the system, and the system meets the accuracy constraint on the dataset, then the constraint is also fulfilled on the data $M$ to be graded, i.e.

\begin{small}
\begin{align*}
&\big( C_{D_M,Diff}^i(T^i,T^*) \geq C_{D_V,Diff}^i(T^i,T^*) \big) \text{\phantom{abcd} Acc. on $D_M$ on difficult samples better than on $D_V$}\\
&  \wedge \big(C_{D_V,Easy}^i(T^i) > C_{min}^i\big) \text{\phantom{abcd}Constraint on accuracy for easy samples fulfilled on $D_V$}\\ 
&\Rightarrow C_{D_M,Easy}^i(T^i) \geq C_{min}^i \text{\phantom{abcd}Constraint on accuracy for easy samples fulfilled on $D_M$}\\ 
\end{align*}
\end{small}

Accuracy on difficult samples tends to be lower than those on easy samples. Thus, we cannot simply check if accuracy $C_{D_M,Diff}^i(T^i,T^*)$ is above the minimum $C_{min}^i$.\footnote{Mathematically speaking, fulfilling this condition is not necessary but sufficient to show that also $C_{D_M,Easy}^i(T^i) >C_{min}^i$.} Furthermore, the above assumption is probabilistic as we shall elaborate on in the evaluation, i.e., the bigger $\Delta:=C_{D_M,Diff}^i(T^i,T^*) - C_{D_V,Diff}^i(T^i,T^*)$ the more likely the conclusion that the constraint is fulfilled holds.
While a positive $\Delta$ (better outcome on the data to grade $D_M$) indicates no issues, the more negative $\Delta$ is, the less the autograder should be trusted. A negative $\Delta$ does not necessarily rule out the use of the autograder even if no compromises on accuracy constraints should be made. A teacher might increase thresholds $C_{min}^i$ and/or $C_{min}^c$ (by $|\Delta|$) which reduces the number of samples judged as easy but increases accuracy. Large $|\Delta|$ can originate from the autograder but also the teacher and statistical variation due to having a small dataset $D_M$. Specifically, the teacher's grading on difficult samples should contain few errors upon validation of the autograder.
%Let the  threshold $T^*$ . 

There is a small subtlety that can happen if any of the minimum accuracy constraints $C_{min}^i$, $C_{min}^c$ is below that of the autograder's average performance as outlined in the last paragraph of Section \ref{sec:defs}. Assume, for example, an autograder achieves a maximum accuracy level of 90\% for the threshold $T^*$, if all samples are graded. Further, assume we aim for $C_{min}^i=100\%$ but only for $C_{min}^c=85\%$.
In this case, it likely holds that $T^c<T^*$. Samples with similarity less than $T^*$ would be graded as incorrect to maximize (overall accuracy). However, the lenient constraint $C_{min}^c$ (and thus $T^c$) allows the autograder to grade samples as ``correct'' that would not be graded as ``correct'' to maximize overall accuracy. In such a case, there are also no samples that are deemed difficult to grade as correct, i.e., the samples in the range $[T^*,T^c]$ are empty since $T^c<T^*$. To verify, a teacher must manually grade a subset of $D_R(T^c,>) \subset D_M(T^c,>)$ that are deemed easy to grade and compare the accuracy on the subset with the constraint $C_{min}^c$, i.e., check if $C_{D_R,Easy}^c(T^c)>C_{min}^c$. Since the test is performed on a subset, statements are probabilistic, i.e., we get a probability $p(C_{D_M,Easy}> C_{min}^c| C_{D_R,Easy}> C_{min}^c)$. The probability can be increased by using a larger subset $|D_R(T^c,>)|$. The same reasoning applies if the constraint on accuracy for incorrectly graded samples is smaller than the system's expected performance.

% \begin{figure}[htbp]
% \includegraphics[width=\linewidth]{validationExp.PNG}
% \caption{Illustration of datasets involved in validation of the autograder. Accuracy $C^{-i,-c}_{T}$ on the manually graded data $D^{-i,-c}_{T}$ should be comparable to the accuracy $C^{-i,-c}_{V}$ on the large validation data $D^{-i,-c}_{V}$} \label{fig:Val}
% \end{figure}

\section{Evaluation}
\subsection{Setup}
To train the baseline, models for LaBSE and the (uncased) multilingualBERT were  taken from the open-source library HuggingFace\cite{wol20}. We used the following hyperparameter setting: the maximal question length was 128 characters (characters exceeding the limit were discarded), the maximal answer length was 64, batch size was 32, freeze was 8 (i.e., the first 8 layers were frozen), the cosine $margin$ used in the loss computation was 0.2. The linear output was 128, the learning rate was 0.000025 and epsilon $\epsilon=1e-8$. We used 10'000 questions (and all its answers) for validation and another 10'000 questions (and all its answers) for testing and the rest for training, i.e., 20'000 questions (and all the answers for those questions) were not included in the training data. Thus, either a question (and all its answers) belonged to the training set or the test set. We have the three datasets commonly found in machine learning: one for training $D_{TR}$, one for validation $D_{VA}$ and one for testing $D_{TE}$. In addition, to evaluate the autograder validation procedure outlined in Section \ref{sec:val}, we consider random samples of 5\% of the dataset $D_V=D_{TE}$ of the test dataset to be used as a sample $D_M$ to be graded by the teacher. We do not estimate any parameter or tune any hyperparameter on $D_V$, i.e., $D_{TE}$. To simulate poor autograder performance, we alter the label of a fraction $f=0.2$ of correct predictions of both the easy and difficult questions so that they appear erroneous. 

We focused on the learning rate, the number of frozen layers and the cosine margin, and the number of neurons in the linear out layer for hyperparameter tuning. Hyperparameter tuning only yielded an accuracy gain of about 1\%, and is therefore not an essential task. %was done on Google Cloud ML costing several thousands of dollars but .
%For the hyperparameter tuning the multilingual BERT model is used. The strategy is to declare some hyperparameters to tune.

\subsection{Results}
\noindent\textbf{Comparison of LabSE and BERT:}
Results are shown in Table~\ref{tab3}, where the threshold $T$ for similarity to distinguish correct from incorrect answers has been chosen to maximize accuracy. Since LaBSE performed better, we will focus on it for the rest of the paper. 
\begin{table}[htbp]
%\caption{Accuracy for LaBSE and BERT depending on the similarity threshold Ts}
\centering
\begin{tabular}{|c|c|c|c|c|}
\hline
\multicolumn{1}{|l|}{\textbf{Model}} & \multicolumn{1}{l|}{\textbf{Class}} & \multicolumn{1}{l|}{\textbf{Precision/Recall}}  & \multicolumn{1}{l|}{\textbf{Accuracy}} & \textbf{Threshold $T$}      \\ \hline
BERT       & 0                                   & 0.84 / 0.83                       & \multirow{2}{*}{0.845}                & \multirow{2}{*}{0.636} \\ \cline{2-3}
(Baseline)  & 1                                   & 0.85    / 0.86     &                                                          &                                                                 \\ \hline
LaBSE       & 0                                   & 0.87   / 0.81                                                                       & \multirow{2}{*}{0.853}                & \multirow{2}{*}{0.608} \\ \cline{2-3}
(Baseline)                     & 1                                   & 0.84 / 0.89                                                       &                                        &                         \\ \hline
BERT          & 0                                   & 0.86   / 0.82                               & \multirow{2}{*}{0.853}                & \multirow{2}{*}{0.754} \\ \cline{2-3}
(Tuned)                               & 1                                   & 0.84      / 0.88                                                                         &                                        &                         \\ \hline
LaBSE         & 0                                   & 0.88   / 0.82                            
                          & \multirow{2}{*}{0.865}                & \multirow{2}{*}{0.753} \\ \cline{2-3}
        (Tuned) & 1                                   & 0.85 / 0.90                                                                       &                                        &                         \\ \hline
\end{tabular}
\caption{ Results for models including hyperparameter tuning.}\label{tab3}
\end{table}

\begin{figure}[htbp]
\centering
\includegraphics[scale=0.4]{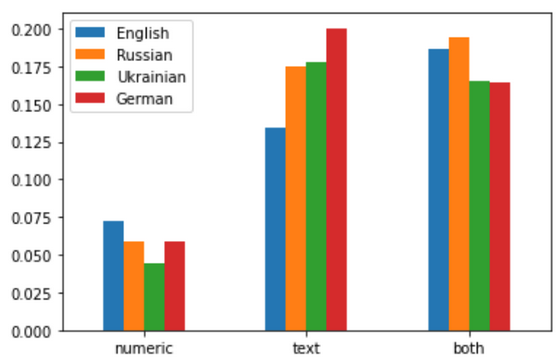}
\caption{Misclassification of autograder given that correct answer is either numeric, textual or contains both elements. } \label{fig9}
\end{figure}

\smallskip

\noindent\textbf{Impact of data characteristics:} We also investigated correctness depending on languages, answer and ground truth length, question type, and available answers per question as shown in Figure \ref{fig9}. We noticed that correctness varies with language and type of question. Generally, grading is most accurate for numeric answers with error rates being less than half of other answer types. This is not unexpected since semantic variation in answers tends to be limited given a single number is expected, e.g., identical answers for ``1'' are, for example, ``1.0'', ``1$\pm$ 0'', ``One'', and ``one''. Drawing conclusions from differences in language must be made with care since the question types employed also vary across languages. Overall differences tend to be less than 8\% between the language with the lowest and highest error. We also investigated the impact of the ground truth length and accuracy of the autograder. We found no strong dependence.% (see Appendix). 

We also conducted a more qualitative assessment by investigating errors of 300 question-answer pairs by the autograder. Interestingly, the errors did not just point to poor grading by the autograder but also by the teacher. There can be two types of annotation errors by a teacher. She can provide the wrong ground truth, i.e., a reply that is an incorrect answer to her own question. She can also error in the comparison, i.e., she alleges that the student's reply does not match the correct reply, although it does. In the investigated question-answers, we found two incorrectly graded answers, i.e., the answer by the student and the ground truth matched exactly, but the answer was still deemed incorrect. Two other graded answers by teachers were debatable, e.g., the answer ``could,need to, can, can't,may'' was deemed incorrect given the correct answer of ``1-could.2-need to.3-can.4-cant.5.-may''.   
Grading among teachers was not consistent, i.e., sometimes questions related to language with typos were considered correct, sometimes not. In some cases, the ground truth contained typos, which might confuse the autograder, e.g. ``true'' was spelled as ``treue'', a German word for ``loyality''. Overall, we believe that the errors do not have a very large impact, i.e., less than proportional. That is, an error rate of $R$ of all annotated question-answer pairs reduces the accuracy of the autograder by at most $R$, since deep learning models have shown surprisingly high levels of robustness to label noise \cite{rol17}.

Common errors by the autograder for language were considering typos or plural forms as correct, e.g. ``a poem was being writen'' or ``man's clothes'' instead of ``men's clothes''. Interestingly, sometimes also the opposite held true, i.e. the autograder considered an answer with a typo as incorrect despite the teacher grading it as correct, e.g. ``unistall'' instead of ``uninstall''.
For math, close answers (sometimes with syntax errors) were commonly deemed correct, e.g. ``$(3y-4)(y^+7)$'' was deemed correct though it should have been ``$(3y-4)(y^2+7)$'' as well as  ``7/80'' though it should have been ``7/40''.
Other questions seemed extremely difficult to grade correctly, since they might have strongly relied on order or they might be fairly unique, e.g., asking for keyboard shortcuts in the ``Adobe'' software  with the answer being ``ctrl+t ctrl+shift+alt ctrl+j ctrl+z ctrl+s''. Overall, even if data quality was higher (no errors in ground truth) and answers were graded more consistently (either a typo is always an error or never), it became apparent that merely relying on semantic similarity might not suffice due to the large number of questions and the fact that an autograder cannot reason logically.

\smallskip
\begin{figure}%
    \centering
    \begin{subfigure}[b]{\textwidth}
    \centering
    \includegraphics[width=0.8\linewidth]{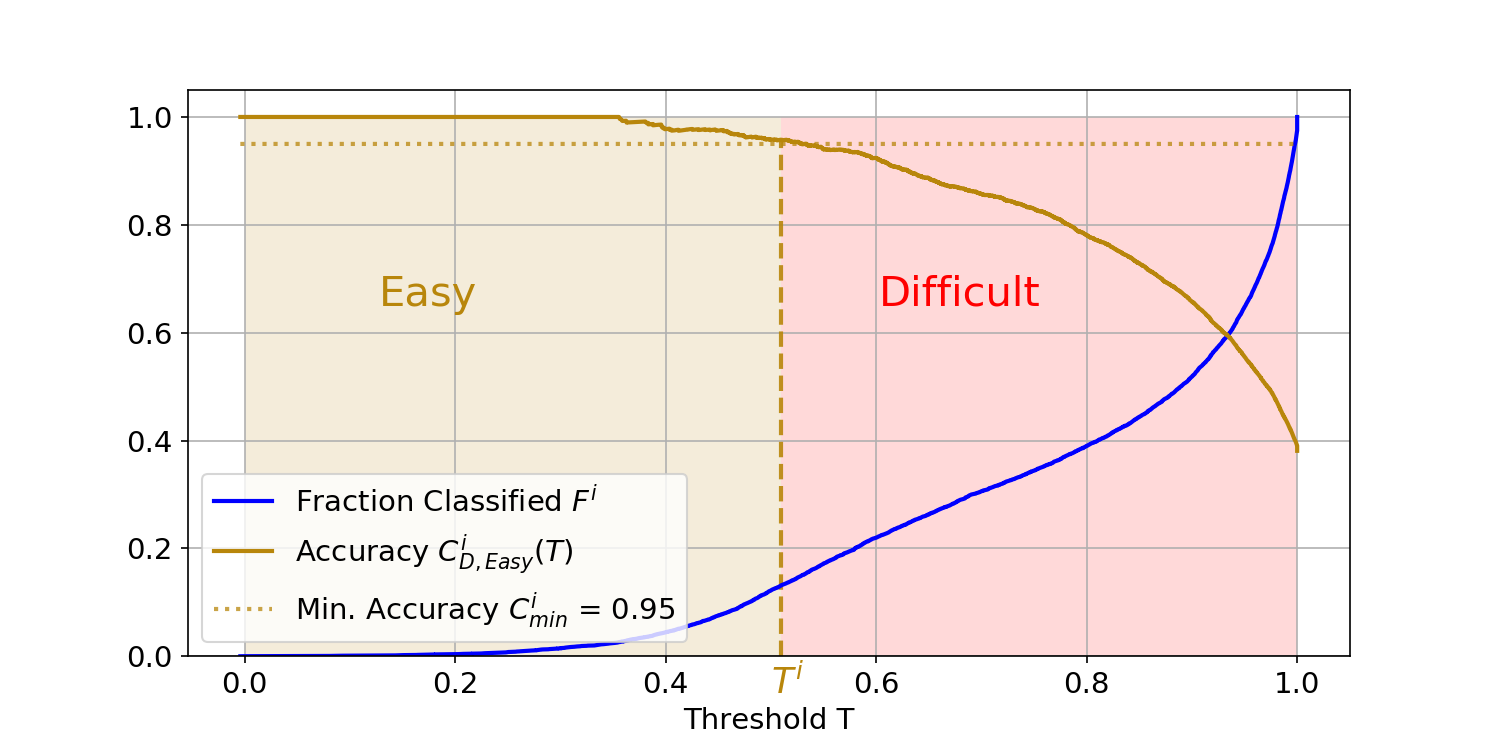} 
    \caption{Accuracy $C^i_{D,Easy}$ for samples classified as incorrect and the fraction of classified samples $F^i$, depending on the threshold $T^i$ for $D=D_{TE}$. }
    \label{fig6}
    \end{subfigure}
    \hfill
    %\quad
    \begin{subfigure}[b]{\textwidth}
    \centering
    {{\includegraphics[width=0.8\linewidth]{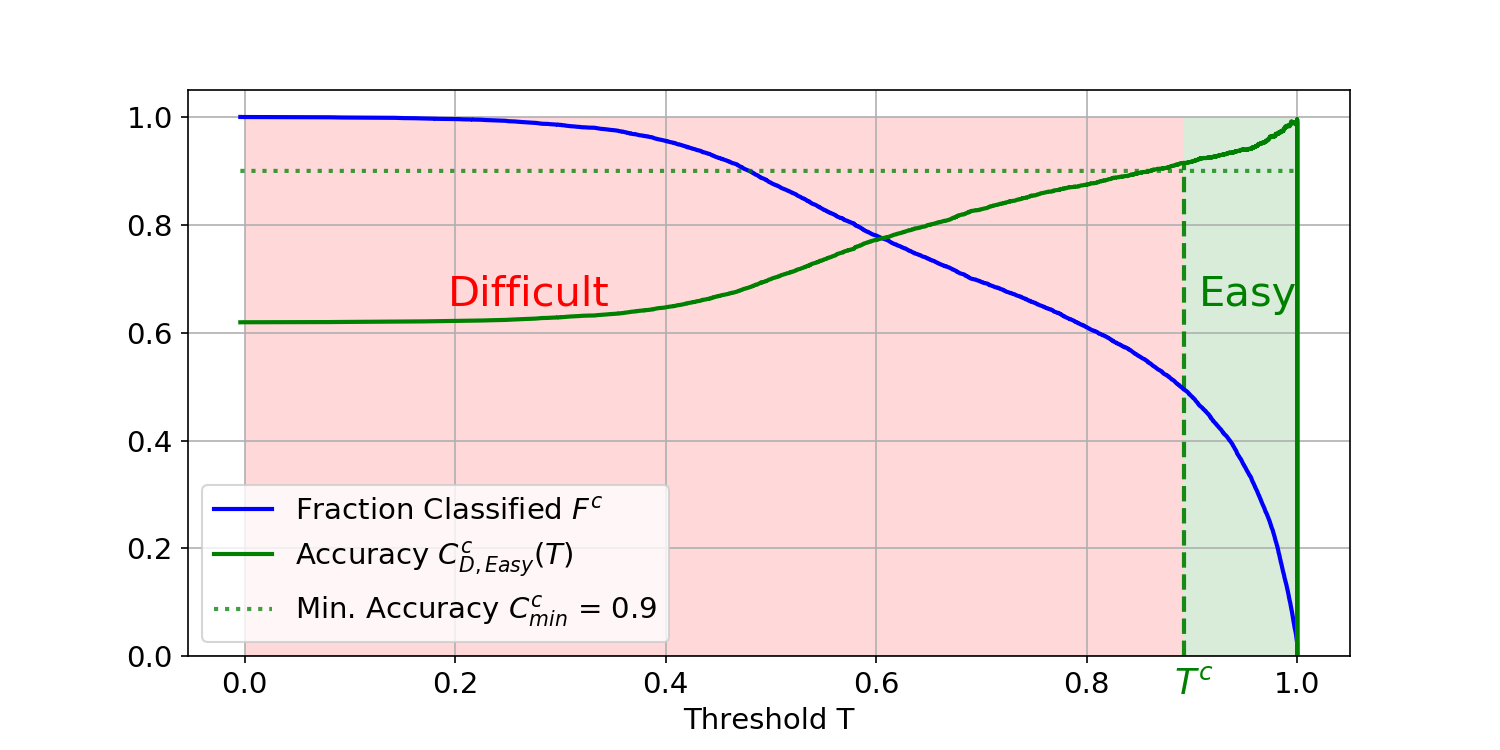} }} 
    \caption{ Accuracy $C^c_{D,Easy}$ for samples classified as correct and the fraction of classified samples $F^c$ depending on the threshold $T^c$ for $D=D_{TE}$.}
    \label{fig7}%
    \end{subfigure}
    \hfill
    %\quad
    \begin{subfigure}[b]{\textwidth}
    \centering
    {{\includegraphics[width=0.8\linewidth]{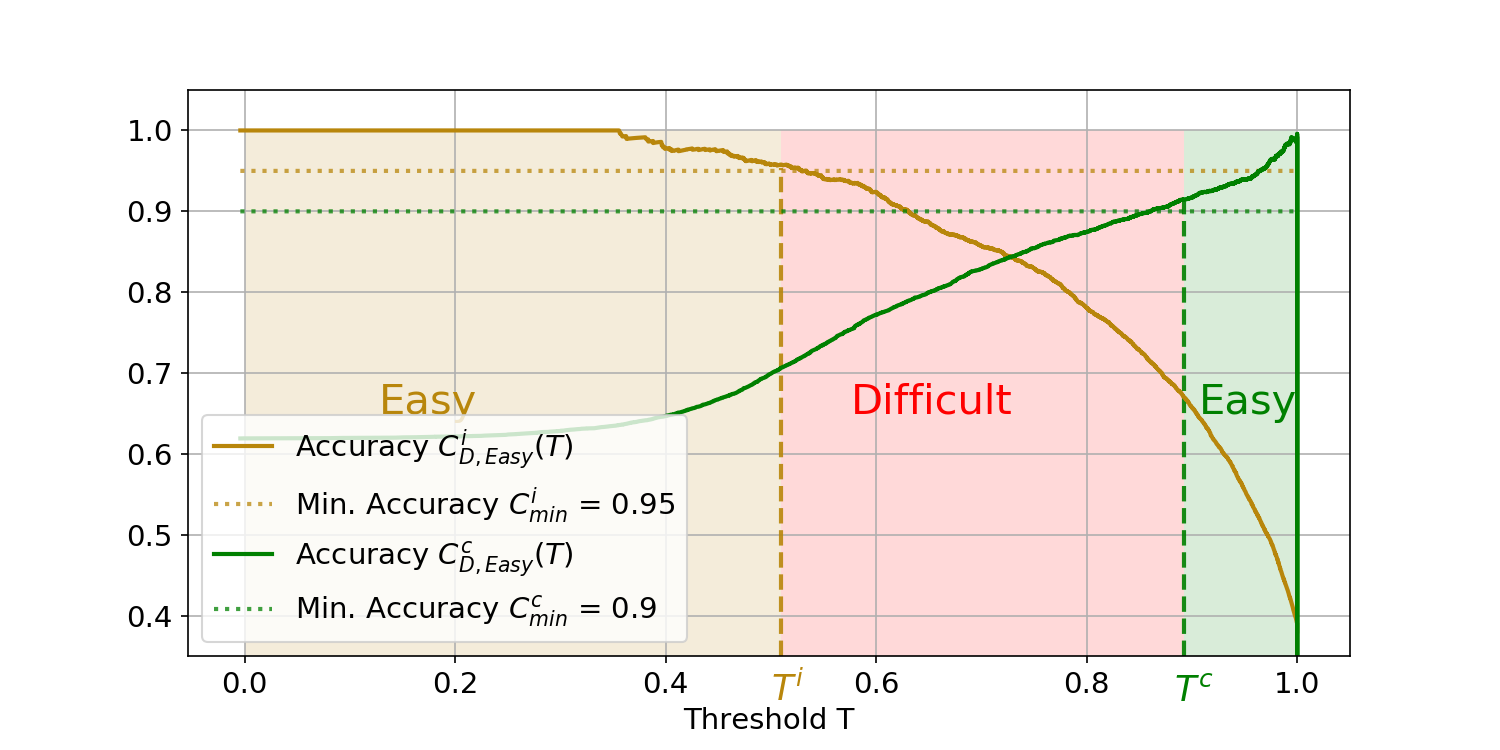} }}
    \caption{ Accuracy $C^c_{D,Easy}$ and $C^i_{D,Easy}$ depending on the thresholds $T^c$ and $T^i$.}
    \label{fig8}
    \end{subfigure}
\caption{ Accuracy $C^c_{D,Easy}$ and $C^i_{D,Easy}$ depending on the thresholds $T^c$ and $T^i$. For single threshold, the best accuracy of about 86.5\% is given by $T^*$. Question-answer pairs with a similarity between 0.5 to 0.89 are deemed difficult to grade, i.e., including them in the classification yields an accuracy $C^c_{D,Easy}$ and $C^i_{D,Easy}$ below the user given thresholds $C_{min}^i$=95\% and $C_{min}^c$=0.9\% }
    
\end{figure}
\noindent\textbf{Results when limiting autograding using thresholds $T^c$ and $T^i$:} Generally, we consider different constraints on the minimum accuracy $C_{min}^i$ and $C_{min}^c$ and we compute the corresponding thresholds $T^c$ and $T^i$ for classification (as introduced in the model section). We estimated $T^c$ and $T^i$ using validation data. We report accuracies $C^c_{D,Easy}$ and $C^i_{D,Easy}$ as well as the fractions of classified samples $F^i$ and $F^c$ on test data $D=D_{TE}$. 
%In particular, autonomous grading of exams requires better performance. Furthermore, a teacher might aim to constrain the type of errors, i.e., avoid judging correct answers as incorrect more than the other way around. an accuracy of 86.5\% when grading all data might not be sufficient for all application cases of an autograder. To improve ac, w

Figures \ref{fig6} and \ref{fig7} illustrate the case, when using either the threshold $T^c$ or $T^i$. Figure~\ref{fig8} shows the scenario using both. In Figure \ref{fig6} above the threshold $T^c$ a sample is deemed ``easy to classify'' and classified as correct. Below the threshold, a sample is considered difficult and it is graded manually. For a minimum accuracy $C^c_{min}$ of 0.9 about 49\% of all samples get classified. In Figure \ref{fig7} a sample is classified by the autograder only if its similarity is below the threshold. In this case, it is classified as incorrect. When demanding an accuracy $C^i_{min}$ for incorrect samples of 95\% only 13\% of all samples get classified. Overall, 62\% of all samples are classified and the overall accuracy of classified samples is 92\%. %, i.e., 95.2\% for the 13\% samples that are classified as incorrect and about 89.8\% for the 49\% classified as correct

An accuracy of about 90\% is deemed insufficient to deploy the system for fully automatic grading of high-stake exams. However, it might be valuable to grade trial exams that a student does for preparation or to point out manually graded questions that differ from the autograding and should be verified. A teacher like any human is subject to mistakes from time to time. She might provide wrong marks “uniformly at random” due to occasional lack of attention. These mistakes are fairly likely discovered. If a teacher achieves an accuracy close to 100\% then a system with an accuracy of 90\% would differ in about 10\% of marked answers, which a teacher should double-check. Thus, by checking 10\% of all marks of the classified samples and resolving any wrong marking, a teacher could discover 90\% of all sloppiness mistakes. Overall, since we classify only about 60\%, a teacher needs to check about 44\%\footnote{The 38\% of non-classified answers and 10\% of the classified answers (62\% got classified), yielding 44\% in total}  of answers. Potentially, this approach could save more than 50\% of work.

Error rates for humans for repetitive tasks are in the range of 0.5\% to 6\% (p. 412 in \cite{smi17}). While such numbers vary from individual to individual we assume an accuracy of 98\% for a teaching assistant grading an exam once.\footnote{High-stake exams are often reviewed by a second assistant leading to higher overall accuracy.} If we use $C_{min}^c=0.98$ and we do not allow the autograder to fail any students, we get a possible workload reduction of slightly above 10\% and overall accuracy of slightly above 98\%.
%Demanding the same accuracy for our model yields that we could classify 13\% of all answers automatically. This is a significant work reduction.

\smallskip
\begin{figure}%
    \centering
        \begin{subfigure}[b]{\textwidth}
        \centering
    \includegraphics[width=0.9\linewidth]{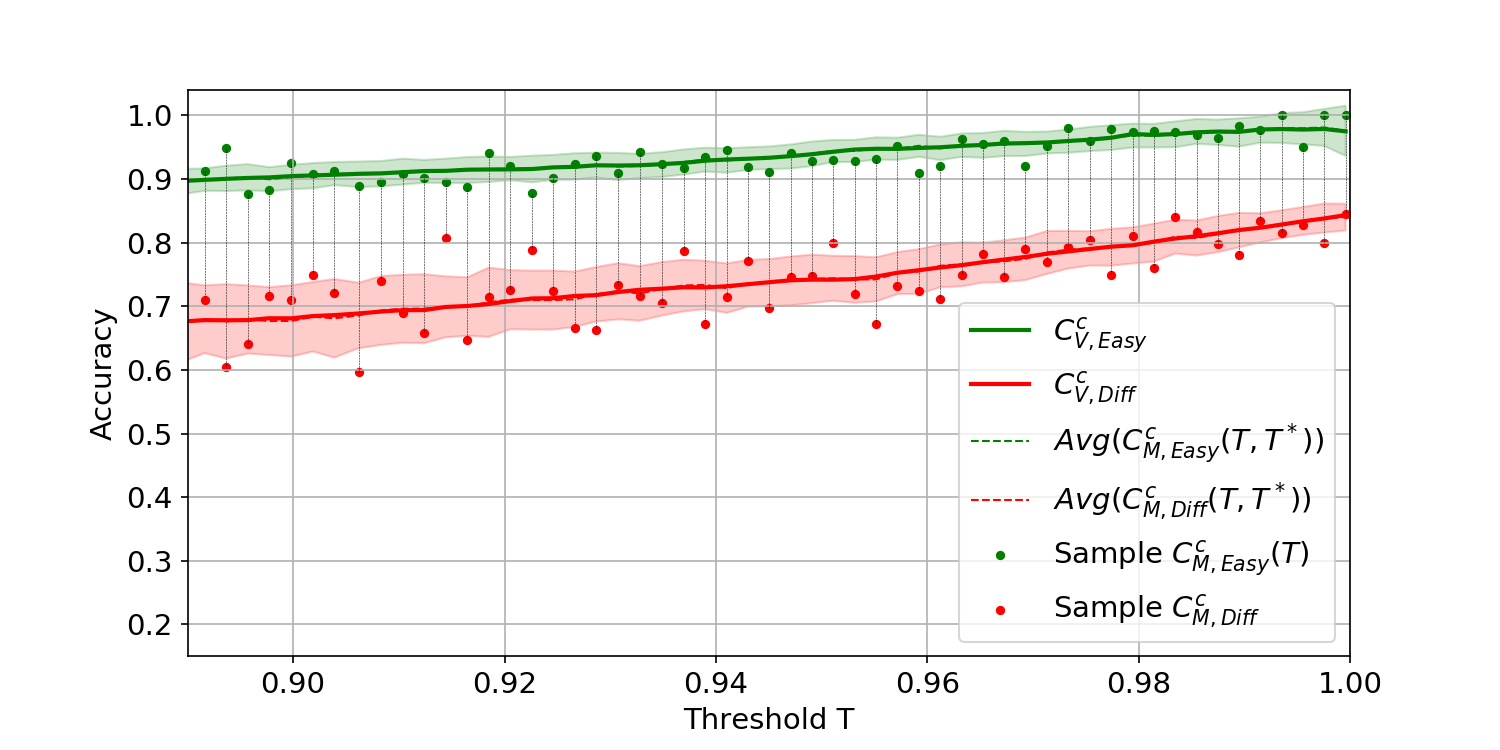} 
    \caption{Normal autograding on $D_M$  }
    \end{subfigure}
    \hfill
        \begin{subfigure}[b]{\textwidth}
        \centering
    \includegraphics[width=0.9\linewidth]{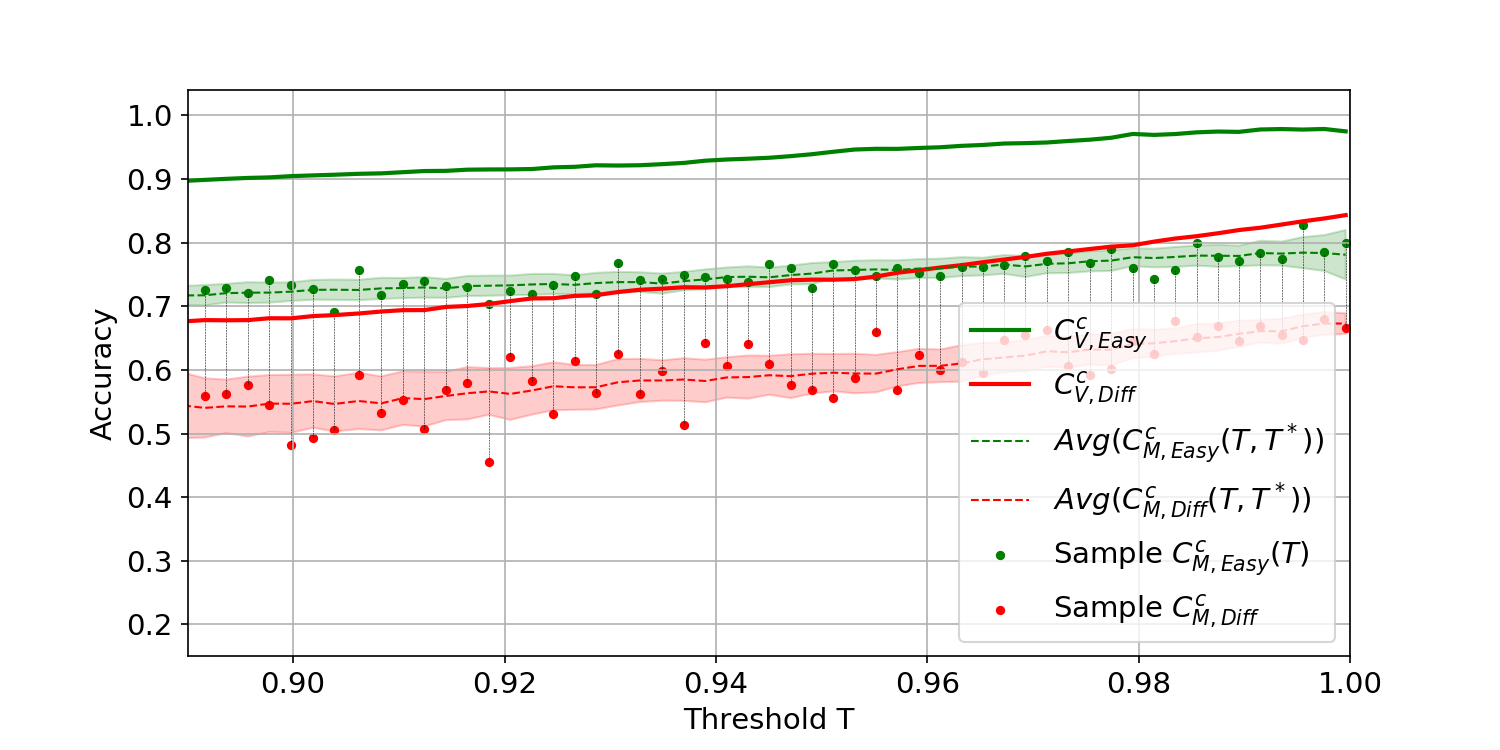} 
    \caption{Poor autograding on $D_M$ }
    \end{subfigure}
\caption{Accuracy dependent on threshold $T^c$ for easy and difficult to grade questions on the large set $D_V$ and small sets a teacher should grade ($D_M$) and aims to validate the autograding. The smaller dataset $D_M$ exhibits higher variation, but overall behaves fairly similarly. For poor autograding we flip 20\% of samples matching the ground truth and classified as correct to incorrect yielding a shift of accuracy.}
    \label{fig:valsamp}%
\end{figure}

\begin{figure}%
        \centering
        \begin{subfigure}[b]{\textwidth}
        \centering
    \includegraphics[width=0.9\linewidth]{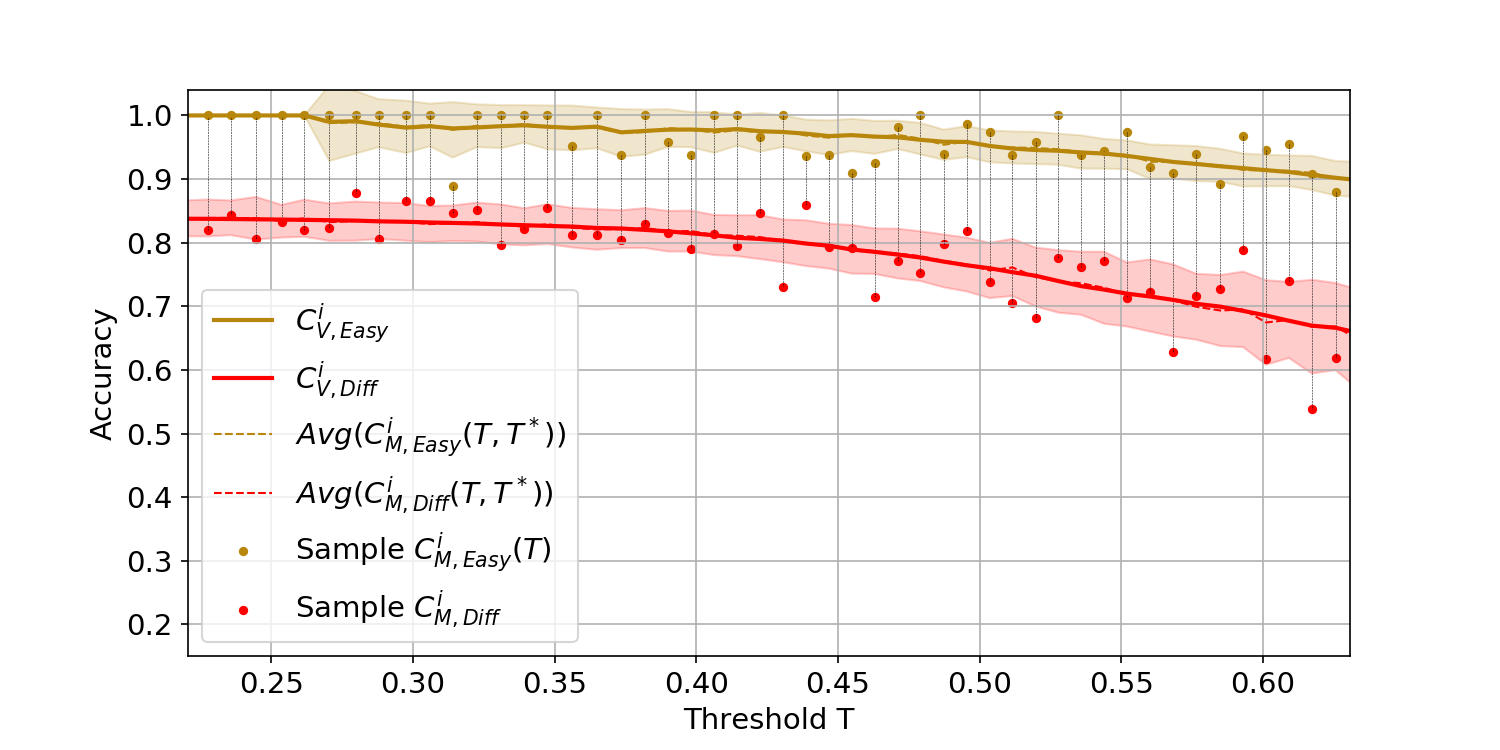} 
    \caption{Normal autograding on $D_M$  }
    \end{subfigure}
    \hfill
        \begin{subfigure}[b]{\textwidth}
        \centering
    \includegraphics[width=0.9\linewidth]{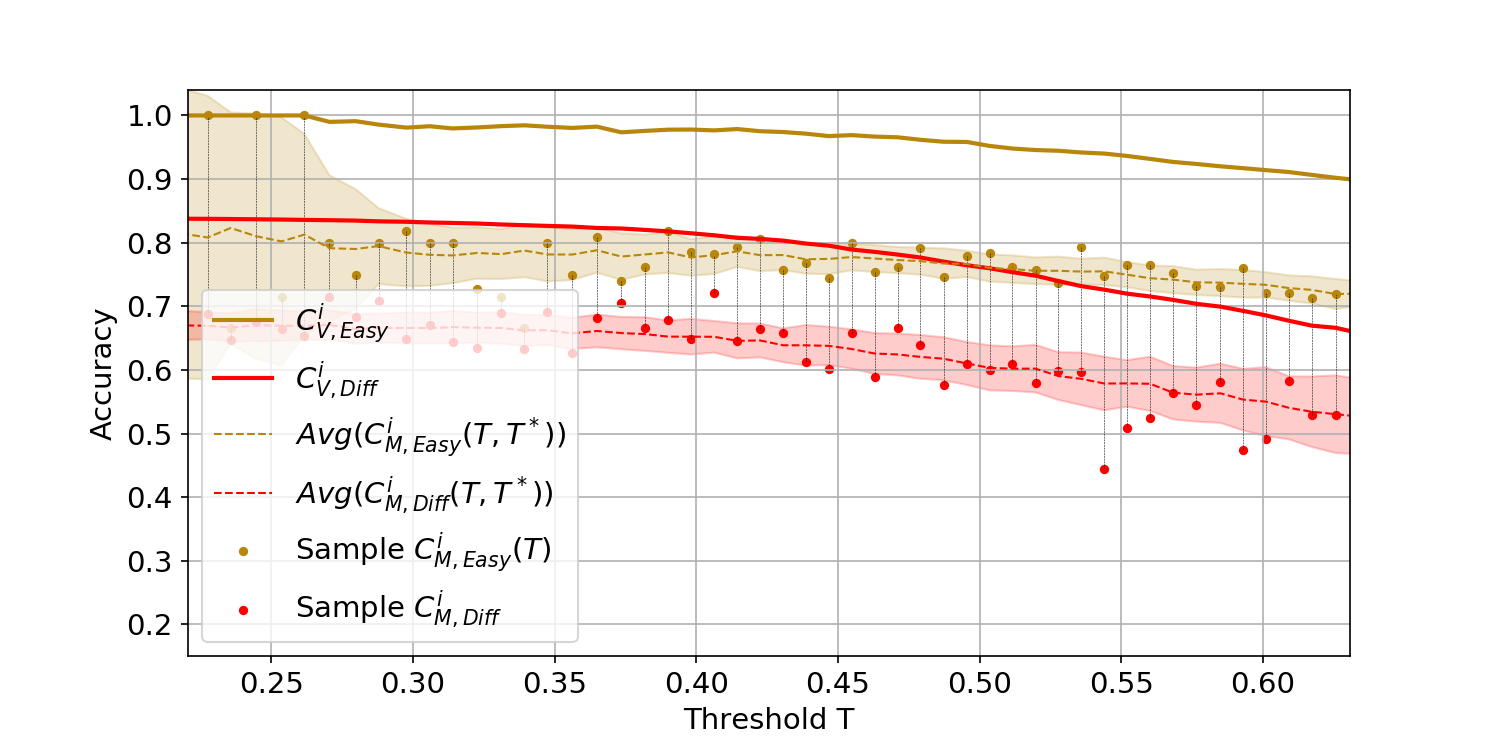} 
    \caption{Poor autograding on $D_M$ }
    \end{subfigure}
\caption{Accuracy dependent on threshold $T^i$ for easy and difficult to grade questions on the large set $D_V$ and small sets a teacher should grade ($D_M$) and aims to validate the autograding. The smaller dataset $D_M$ exhibits higher variation but overall behaves fairly similarly. For poor autograding we flip 20\% of samples matching the ground truth and classified as correct to incorrect. Accuracy gets lower in a similar fashion for easy and difficult to grade questions.}
    \label{fig:valsamp2}%
\end{figure}
\noindent\textbf{Results for Validation of Autograder:} 
%For the dataset $D_V$ overall accuracy is maximized for a threshold $T=0.76$, i.e., we should classify samples less than $T$ as incorrect and those above as correct.
For the dataset $D_V$ overall accuracy is maximized for a threshold $T^*=0.76$ leading to an accuracy of $C^*=86.5\%$.
%(corresponding to the for $C_{min}^c\in [0.9,1]$ for $C_{min}^i \in [0.9,1]$
We show accuracy for easy and difficult subsets $D_M$ as well as the large set $D_V$ depending on varying thresholds in Figure \ref{fig:valsamp} and in Figure \ref{fig:valsamp2}. We discuss the behavior of accuracy on easy and difficult samples being subsets of the large dataset $D_V$ first. The (average) performance gap between the grading of easy and difficult samples is more than 10\%, showing that difficult samples are harder to grade for the autograder.

The gap is non-constant. It is larger the closer the threshold $T^c$ or $T^i$ is to $T^*$. For samples near $T^*$ predictions are most uncertain. The accuracy improves faster than linear when moving away from $T^*$. 

We observe large variance in accuracy for individual, randomly chosen subsets $D_M \subset D_V$ for both easy and difficult samples. Overall, variance is larger for  accuracy computed on difficult samples than on easy samples. The reasoning is the same as before, i.e., uncertainty is higher on difficult samples, leading to larger variance.

Furthermore, variances are non-constant, i.e., both easy and difficult samples are  heteroscedastic. The variance on difficult samples is largest on samples near $T^*$, and for easy samples, it is lowest near $T^*$. This follows since the closer the threshold to $T^*$, the fewer difficult samples exist, and the more easy ones and vice versa. Variance in accuracy estimates decreases with the number of samples used.  Concretely, the large variance for difficult samples $D_M$ (Figure \ref{fig:valsamp2}) stems from the fact that there are only about 5 samples with such a low similarity score. Thus, if we alter just a single sample, accuracy can change by 20\%, leading to large variance.

%From a statistical perspective, one might introduce a lower bound of samples that should be in a set to estimate the accuracy. 
Furthermore, it is beneficial if deviations for the difficult and easy samples on $D_M$ from $D_V$ correlate strongly, i.e., it could be said that if the performance on difficult samples is below that on $D_V$ by $t$ it is also below the same amount $t$ for easy samples. Unfortunately, correlation is only weak, as indicated by the varying length of vertical lines connecting outcomes for easy and difficult samples in Figures \ref{fig:valsamp} and \ref{fig:valsamp2}. It can be concluded from visual inspection and a Shapiro-Wilk test that for a large range of thresholds $T^c$ accuracy on easy and difficult samples for a randomly chosen set $D_M$ is distributed roughly normal, i.e., $C^c_{M,Easy}(T^c) \sim N(C^c_{V,Easy}(T^c),\sigma(T^c)$ and $C^c_{M,Diff}(T^c) \sim N(C^c_{V,Diff}(T^c),\sigma(T^c)$. Analogous thoughts apply for $T^i$. Exceptions are caused due to skew (accuracy is at most 1) and due to very small sample sizes. Skew is most prevalent for $T^c$ close to 1 (or small $T^i$). Sample sizes are also small for such thresholds, e.g., few samples have a similarity beyond $0.999$ or below $0.2$.

To summarize, a deviation of a few percent in accuracy between the large dataset $D_V$ and a small dataset $D_M$ to grade are common. While this might be for the better (higher accuracy), it could also be for the worse (lower accuracy). Such deviations are also expected due to the statistical nature of machine learning. They can also be caused by the teacher, e.g., if the difficult samples are graded much more poorly than on average. A teacher can use more stringent constraints ${C'_{min}}^c>C_{min}^c$ than her desired accuracy to reduce the risk that actual performance is below $C_{min}^c$. On the positive side deviations are mostly only a few percent. For large autograder errors on $D_M$, i.e., if we change 20\% of difficult and easy samples of $D_M$ to be wrongly graded, we see that this results in a drastic shift of accuracy for both types of questions, i.e., those that are easy and difficult to grade. In practice, the essential questions are: \\
\noindent (i) ``What is the risk that an accuracy constraint $C_{min}^c$ is violated?''  \\
\noindent(ii) ``How much is a teacher's workload reduced?''\\
More technically, Question (i) can be posed as:
What is the risk the accuracy on easy samples $C^c_{D_M,Easy}$ is lower by $m$ or more compared to the provided accuracy $C_{min}^c$, i.e., $C^c_{D_M,Easy}+m < C_{min}^c$?  In that light, Question (ii) becomes: How often is the autograder still employed for such an $m$?'' 
% given that $C^c_{V,Diff}-C^c_{M,Diff}\geq k$
%Here, $k$ is a parameter that allows to control the risk, i.e., the smaller $k$ the lower the risk.

In our case, we use $m=0$, i.e., we only accept autograded answers if the accuracy on the difficult samples in the small dataset is at least that on the large dataset. For $m=0$, we obtain a risk of 50\% that the constraint of $C_{min}^c$ is violated. At the same time, we also only use the autograder half the time, i.e., for 50\% of all the datasets $D_M$. Assume we aim to avoid the scenario we deemed ``poor autograding'' with a performance deterioration of $m=0.2$.  Then, we only use the autograder if the following two condition apply: (i) the constraint $C^c_{V,Diff}-C^c_{M,Diff}>0$ is fulfilled and (ii) easy samples are graded very poorly, i.e., $C^c_{M,Easy}+m < C_{min}^c$. From Figure \ref{fig:valsamp2} we see that the probability for (i) is very low, i.e., a sample $C^c_{D_M,Diff}$ would roughly have to deviate about 2.5-5 standard deviations. For a normal distribution, this corresponds to a probability between 0.006 and $3\cdot 10^-7$. 
%accuracy on difficult samples is above that a sample is about and $m=0.2$, we obtain a risk of less than 1\% (on average), while using the autograder for 85\% of all cases. 
In our concrete scenario demanding an accuracy level of $C_{min}^c=98\%$ and $C_{min}^i=100\%$, the risk is below 0.0001. Furthermore, even if the autograder grades all answers, i.e., 100\%, incorrectly, the overall error rate would increase by about 10\% since the autograder only grades easy samples, which amounts to about 10\% of all samples.

\section{Discussion and Limitations}
Autograding has gained increased attention in recent years due to staggering improvements in NLP. However, error rates are typically still around 10\% (aligned with our outcome). While regulatory bodies have been asking for human agency and oversight for AI systems as a guiding principle \cite{eur20}, most research has paid little or no attention to meeting such demands when using and evaluating AI-based autograders.

For automatic grading of high-stake exams, a 10\% error rate is certainly not acceptable. An individual decision of such a system cannot be trusted (to be correct).  We argue that at least human-level performance should be aimed for in general. By delegating work to humans, this is possible. Even if autograders achieve an accuracy comparable to humans on average, they might exhibit stronger variation  and different forms of errors. For example, autograders do not make sloppiness errors due to being tired as teachers might, but they might make decisions in an unethical way. Letting a student fail who should pass might be worse than letting a student pass who should fail. Furthermore, our empirical investigation highlighted that accuracy varies to some degree with question types and other factors such as language. Generally, there are concerns with respect to concept drift, i.e., a mismatch of training and test data that might lead to incorrect predictions in addition to biases. This creates the need for the validation of autograders through manual grading. However, manual work is a major hurdle and significantly reduces the benefits of autograding. While our work aimed at minimizing this effort, manual grading of a significant portion of answers is still required.

We believe that even minor biases resulting from exposure to incorrect predictions should be avoided. However, our results indicate that only large deviations from  expectation can be detected reliably on small datasets. Identifying if an autograder shows (on average) minor forms of bias, i.e., grading better or lower than its specification, requires the union of many exams, i.e., it can only be found out after monitoring a teacher for a prolonged amount of time.

While this might sound unsatisfactory, it should be kept in mind that, in practice, for non-automated grading little monitoring takes place since it is very labor intense. At the same time, it is known that teacher's grading suffers from multiple shortcomings beyond simple ``sloppiness errors'' such as ``rater drift'' over time or a tendency to be too lenient or too extreme \cite{hos01}; in practice, little is done to address these concerns.

Our work considered only two possible markings: correct or incorrect. In practice, complex questions are often graded gradually using points. We believe that this changes the problem and the value of an autograder considerably. When fine-grained decision-making is required by introducing points, the system might more commonly misjudge (slightly). Essentially the number of outcomes for a question changes from two (correct/incorrect) to the number of possible points, e.g., if points between [0-10] (including 0 and 10) are obtainable, there are 11 outcomes. We believe that in this case, a system using similar NLP techniques like this and other works, will commonly be off by one or more points. If questions are graded by the system in actual exams, a teacher must still double-check the grading, as errors should be avoided. This significantly reduces the system's value though it still has its merits.

Furthermore, a teacher performs multiple tasks beyond grading when examining human works. For example, she might look out for all kinds of cheating, such as plagiarism or even discover psychological distress of students in essays.  Identifying cheating might increase in relevance since AI has also been subject to subtle attacks, where inputs were changed slightly, altering classification behavior drastically. At the same time, such modifications of inputs were inconceivable for a human.  These so-called adversarial attacks have been successful in various areas such as computer vision \cite{akh18} and NLP \cite{zha20} including short answer grading \cite{fil20}.  Therefore, additional tasks performed by a teacher should not be forgotten, e.g., covered by additional tools or integrated into existing tools based on further research.

Additionally, our model could be enhanced by multiple techniques found in other works such as adding domain-specific features \cite{uto20,Sultan2016} or recognizing question types or their difficulty \cite{pad17}, and utilizing specialized methods for each of them though on the downside this increases system complexity. It is especially interesting to leverage rubrics as defined by teachers. While our work aims to establish a methodology that ensures high reliability of autograders, evaluated with real-world data, it is also interesting to assess the proposed process (built into an actual autograding system) and other potential benefits aside from time-saving based on a user study. For example, it might be interesting to see to what extent teachers' satisfaction increases by focusing the grading primarily on more difficult to grade answers.

\section{Conclusions}
Autograding is becoming more and more feasible due to increased data and improvements in NLP. Our evaluation based on fine-tuning state-of-the-art transformers aligns with other works and their conclusion that current performance, i.e., accuracy, cannot live up to human performance in general. However, when difficult to classify samples are deferred to humans, decision accuracy can be raised substantially. Still, classifier accuracy should always be assessed explicitly for a set of novel question-answer pairs, since due to a large diversity of questions, there might be considerable differences in performance on training, validation, and test data.  In future work, we aim to improve models, e.g., by utilizing additional information such as response times of students, implementing feedback to students, e.g., in the form of personalized explanations of autograder decisions \cite{sch19,sch21} and human-to-AI coaches \cite{sch20}, improving trust issues \cite{diet15,dzi03}, and including detection of dishonest behavior.% ersonalize models \cite{sch21,sch18} .and integrate other functions such as plagiarism detection .

%\newpage
%\section*{References}
%\nocite{*}
%\bibliographystyle{IEEEtran}

% \section{Statements and Declarations}
% The authors declare that no funds, grants, or other support were received during the preparation of this manuscript.\\

% \noindent The authors have no relevant financial or non-financial interests to disclose.\\

% \noindent Robin Richner and Micha Riser developed the autograder models. Micha Riser also supported getting and cleaning data. Johannes Schneider developed the main contribution, i.e., the approach shown in Figure 1 title ``Overview of approach'' and detailed in the manuscript. Robin Richner also contributed significantly to the writeup, but most writing was done by Johannes Schneider. Micha Riser reviewed the manuscript. 

%%%\bibliographystyle{spbasic}
\bibliographystyle{spmpsci}
\bibliography{refs.bib}
%%\bibliographystyle{apa}

% \section{Appendix}
% Figure \ref{fig10} highlights that accuracy of the autograder is not strongly dependent on the expected answer length, i.e., the number of tokens in the ground truth answer. A token results from splitting the answer by spaces and punctuation.
% \begin{figure}[htbp]
% \includegraphics[width=0.8\linewidth]{ntokens_anchor_all_types.png}
% \caption{Accuracy of autograder based on number of tokens in ground truth answer.}
% \label{fig10}
% \end{figure}

% Another categorization based on answers including numeric answers as well as focusing on textual answers only is stated in Figure \ref{fig3b}.

% \begin{figure}[htbp]
% \includegraphics[width=0.7\linewidth]{three_types_all_data.png}
% \caption{Distribution of numeric, non-numeric, and mixed answers  } %A ``token'' can be either a number or a word.
% \label{fig3b}
% \end{figure}

\end{document}